\documentclass{article}

\PassOptionsToPackage{numbers,square}{natbib}
\usepackage[preprint]{neurips_2025}

\usepackage[utf8]{inputenc}
\usepackage[T1]{fontenc}
\usepackage{url}
\usepackage{booktabs}
\usepackage{amsmath}
\usepackage{microtype}
\usepackage{xcolor}
\usepackage{graphicx}
\usepackage{array}
\usepackage{tabularx}
\usepackage{enumitem}
\usepackage{float}
\usepackage{fvextra}
\usepackage[hidelinks]{hyperref}

\newcolumntype{L}[1]{>{\raggedright\arraybackslash}p{#1}}
\newcolumntype{Y}{>{\raggedright\arraybackslash}X}
\newcolumntype{P}[1]{>{\raggedright\arraybackslash\hsize=#1\hsize\linewidth=\hsize}X}

\graphicspath{{figures/}}
\DeclareUnicodeCharacter{2212}{\ensuremath{-}}

\title{TrialJev: Open-Weight Single-Pass Eligibility Scoring for Patient--Trial Screening}

\author{%
  Omar Ibrahim\thanks{Corresponding author: \texttt{omar.ibrahim2@uhn.ca}}\\
  AI Hub, University Health Network (UHN)\\
  Toronto, ON\\
  \And
  Grant Hamblin\\
  AI Hub, University Health Network (UHN)\\
  Toronto, ON\\
  \AND
  Benjamin Haibe-Kains\\
  AI Hub, University Health Network (UHN), Toronto, ON, Canada\\
  Department of Medical Biophysics, University of Toronto, Toronto, ON, Canada\\
  Vector Institute for Artificial Intelligence, Toronto, ON, Canada\\
  Structural Genomics Consortium, University Health Network, Toronto, ON, Canada\\
}

\begin{document}
\maketitle

\begin{abstract}
Clinical trial screening requires assessing large numbers of patient--trial pairs to prioritize candidates for detailed eligibility review, often under restrictions on sharing patient records. We introduce TrialJev, an open-weight eligibility scorer that uses LoRA-adapted Qwen models and a pointer head to score natural-language answer options in a single forward pass, without autoregressive decoding. Trained on TREC Clinical Trials 2021 and evaluated on 6,975 eligible-or-excluded pairs from 50 unseen TREC 2022 synthetic patients, TrialJev achieves 80.97\% (Qwen3.5-27B) and 81.84\% (Qwen3.8-27B) accuracy, compared with 79.87\% and 79.93\% for LLMs fine-tuned on the same data to generate verdicts. In matched batch-1 measurements, verdict generation takes 2.45--2.70 times as long as TrialJev's one-pass inference at 9B--27B. Under a common development-fitted decision correction, the two 27B TrialJev models reach 80.99\% and 81.88\% accuracy, compared with 81.28\% for the proprietary hosted Jev API; paired confidence intervals include zero. TrialJev runs locally on a single GPU, requiring approximately 178\,ms per pair at 27B, without transmitting patient text to an external model API. An A/B-logit ablation shows that avoiding decoding, rather than the pointer head alone, accounts for the measured efficiency advantage. TrialJev is intended for first-pass prioritization of candidate trials, complementing rather than replacing explanatory models and clinical review. Evaluation is limited to synthetic patients and implementation-specific timing, and clinical use has not been validated. Adapters, pointer heads, and model cards are released on Hugging Face.

\end{abstract}

%======================================================================
\section{Introduction}\label{sec:1}

Patient-trial matching combines two different demands: identifying promising trials efficiently and establishing why a patient may qualify. Eligibility criteria contain numeric thresholds, negation, temporal constraints and nested clinical conditions. A coordinator must reconcile these with the available patient record, often across many candidate trials. In the TREC 2022 test set a patient has between 14 and 398 judged candidate trials (median 116). TrialGPT demonstrates the value of language-model assistance for this process \cite{ref1}; the TREC Clinical Trials track \cite{ref2} and subsequent matching systems \cite{ref3} provide benchmarks and methods for studying it.

These two demands need different outputs. Establishing why a patient may qualify calls for criterion-level judgments and written explanations, which generative systems provide~[1]. Identifying promising trials efficiently calls for a first pass that gives every candidate pair one score: cheap enough to apply to all of them, valid for every input, so that candidates can be ranked, cut off and routed to review, and ideally computed locally because patient records often should not leave the hospital. Writing a verdict for every candidate serves this poorly. It requires sequential decoding and verdict extraction, the decoded verdict itself carries no probability, and the written answer can fail to parse. In our measurements (Section 5.2), generating a short verdict took 1.5 to 2.7 times as long as direct scoring, while a zero-shot explain-then-answer baseline required 7--25 seconds per pair. This raises a practical question: how does scoring eligibility directly, rather than generating a written verdict, affect accuracy and inference latency?

A single-pass scorer avoids token decoding. The open-weight MedJev \cite{ref4} and kev \cite{ref5} releases implement Jev-style typed decisions \cite{ref6}\dag{} using a Qwen backbone, a LoRA adapter and a pointer head over answer-option spans. TrialJev adapts this design to patient--trial eligibility. It returns a normalized score for each option from the model's hidden states. The underlying architecture also supports multiple questions sharing a patient prefix; our primary experiment uses one binary question per trial, so its measured efficiency comes from the one-pass scoring path, not from demonstrated prefix-cache savings.

The two approaches are complementary rather than competing. A typed-decision scorer prioritizes: it scores every candidate in one pass and returns a normalized score for ranking and cutoffs. Generative and reasoning systems explain: they justify a decision criterion by criterion, which is worth the cost for shortlisted or borderline candidates. In a staged workflow, a retriever proposes candidates, TrialJev prioritizes them, and a clinician or downstream explainer reviews the highest-ranked ones (Section~6). This paper evaluates only the scoring stage; we did not evaluate the complete workflow, explanation quality or staff time.

We evaluate the deployment trade-off first against a genuine generative alternative, a fine-tuned LLM: a LoRA-adapted model trained with full-vocabulary cross-entropy to write ``Eligible'' or ``Excluded'' and stop, on the same data and recipe. We then compare TrialJev with zero-shot prompted LLMs (one word, and explain-then-answer), with untrained base models, and with the hosted Jev API. Separately, an A/B-logit classifier provides an ablation of the pointer readout, using the language model's own output layer and a loss restricted to two label-token logits \cite{ref7,ref8}. This control is a classifier, not a conventionally fine-tuned generative model. Keeping these systems distinct lets us examine both the cost of writing a verdict and whether a pointer head is needed for the observed accuracy.

The contributions are:

\begin{enumerate}
  \item \textbf{An open-weight, single-pass eligibility scorer.} TrialJev adapts the typed-decision design to patient--trial screening across four Qwen backbones, using a cross-year TREC split. Task-trained adapters and pointer heads for every reported configuration and seed are released on Hugging Face.
  \item \textbf{A controlled comparison of scoring and verdict generation.} With the same training data, TrialJev has similar observed accuracy to a conventionally fine-tuned LLM, whose median batch-1 inference latency is 2.45--2.70 times TrialJev's at 9B--27B in the matched-pair implementation. We report paired uncertainty and the prompt, objective and readout differences rather than attributing every gap to architecture.
  \item \textbf{A local alternative to a hosted decision API.} Both 27B TrialJev models have competitive observed accuracy against hosted Jev under a common development-fitted correction. Hosted request latency is reported separately from local inference timing. Hosted responses also differed on repeated calls (1,384 of 2,638 repeated pairs), whereas local pinned weights fix the model version. Local deployment can avoid external model-API transmission; privacy and clinical governance remain separate requirements.
  \item \textbf{Readout and initialization analysis.} The A/B-logit control separates single-pass scoring from the pointer-head design. Clinical-extraction initialization improves the 0.8B pointer model under the tested recipe, while fresh conventional generative fine-tuning succeeds at this size; initialization is not a general prerequisite for learning the task.
\end{enumerate}

The task is binary: eligible versus excluded for pairs that assessors had already judged relevant. It is not retrieval from a registry, not the three-class TREC task, and not an evaluation of clinical use.

\begin{figure}[htbp]
  \centering
  \includegraphics[width=\linewidth]{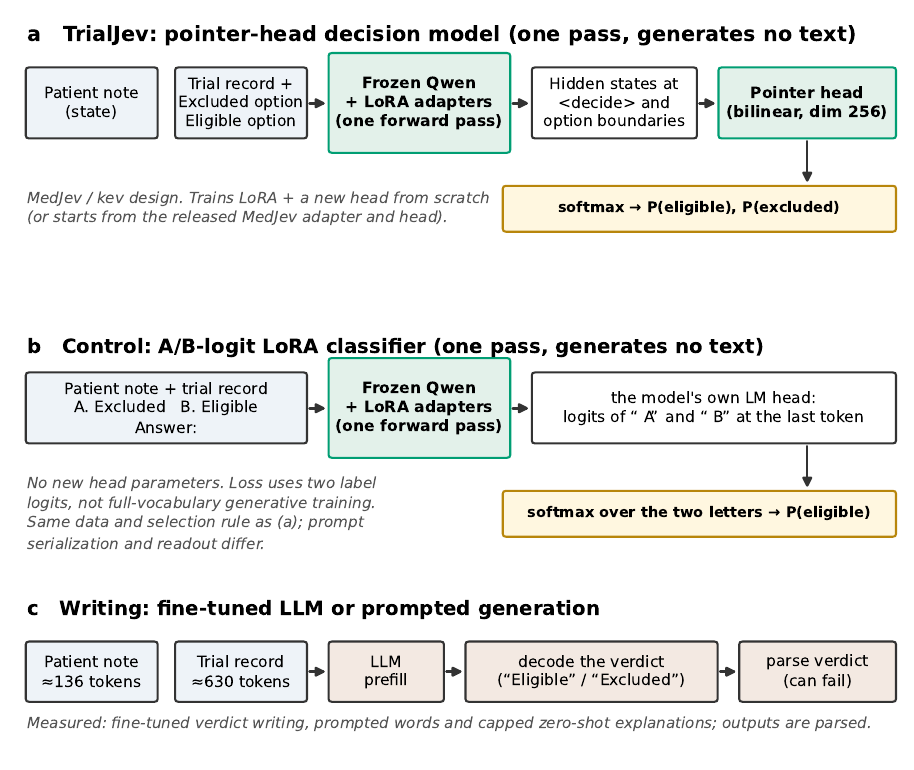}
  \renewcommand{\thefigure}{1}
  \caption{Scoring and writing are distinct inference routes. (a) TrialJev scores option spans from hidden states in one pass. (b) The A/B-logit control scores two label tokens, with training restricted to those logits. Both use the same backbone, data and optimization recipe, but differ in readout and prompt serialization. (c) Generation prefills, decodes and parses a verdict. Our fine-tuned LLM follows this route with an adapter trained to write the answer; the prompted baselines use no adapter. Zero-shot explanations were generated and timed separately.}
  \label{fig:1}
\end{figure}

%======================================================================
\section{Related work}\label{sec:2}

\textit{Generative systems.} TrialGPT \cite{ref1} uses large language models for retrieval, criterion-level matching with explanations, and ranking. Its headline accuracy of 87.3\% is measured per criterion on 1,015 manually reviewed pairs, which is not pair-level eligibility. TrialMatchAI \cite{ref9} combines retrieval with criterion-level reasoning by fine-tuned open models. A system with controlled LLM reasoning over sets of criteria \cite{ref10} reports nDCG@10 of 0.693 on TREC 2022, and SatIR \cite{ref11} handles retrieval by constraint satisfaction and would sit upstream of a scorer like ours. Panacea \cite{ref12} is a foundation model for trial search and patient--trial matching.

\textit{Closest to our task.} VERDICT \cite{ref13} judges eligible versus ineligible on 363 pairs from TREC 2021 and reports F1 of 0.828 with GPT-5-mini and 0.829 with a distilled Qwen2.5-7B formalizer, together with formal decision traces. LLM-Match \cite{ref14} fine-tunes Llama-3-8B-Instruct with retrieval and a classification head and reports AUROC of 0.8102 on TREC 2022, but it merges excluded and eligible into one class against irrelevant, so its task is relevance rather than the eligible-versus-excluded question studied here.

\textit{Typed-decision models.} Jev \cite{ref6}\dag{} is described as returning typed probabilistic decisions about a shared state in one parallel pass. kev \cite{ref5} reproduces the design in open weights on Qwen3.5 with a rank-16 LoRA and a pointer readout, and MedJev \cite{ref4} applies it to clinical-variable extraction. Both are software releases. They motivate our adaptation to eligibility screening and the separate ablation of the pointer readout against language-model logits. Appendix F places these systems in a design-space table.

\textit{TREC.} The 2021 and 2022 Clinical Trials tracks pose ranked retrieval over a ClinicalTrials.gov snapshot with three judgments: not relevant, excluded and eligible \cite{ref2,ref15}. The 2022 track offered the 2021 judgments for tuning, which matches our cross-year split.

%======================================================================
\section{Task and data}\label{sec:3}

For a patient description \textit{p} and a trial record \textit{t}, the model outputs P(eligible | p, t). Labels come from TREC grades: eligible (grade 2) is the positive class and excluded (grade 1) the negative class. Not-relevant pairs are not used, and a trial without a judgment is never treated as a negative. The label therefore means ``excluded from an already relevant trial'', which makes this a second-stage question.

Training and development data come from TREC Clinical Trials 2021: 75 synthetic patient cases written like an admission note \cite{ref2}, with 35,832 judgments (5,570 eligible, 6,019 excluded, 24,243 not relevant). Cases are split by patient, never by pair (seed 13), into 60 training and 15 development patients. The test data come from TREC 2022: 50 synthetic cases, whose 28,419 not-relevant pairs are left out. All 6,975 eligible and excluded pairs are used, without sampling or balancing; they are exactly the 3,939 eligible and 3,036 excluded judgments in the track overview \cite{ref15}. A patient has between 14 and 398 pairs (median 116).

\begin{table}[htbp]
  \centering
  \renewcommand{\thetable}{1}
  \caption{Datasets. Always predicting ``eligible'' scores 56.47\% on the test set.}
  \label{tab:1}
  \footnotesize
  \begin{tabularx}{\linewidth}{P{0.89}P{0.80}P{0.37}P{0.56}P{1.31}P{2.07}}
    \toprule
    \textbf{Split} & \textbf{Source} & \textbf{Pairs} & \textbf{Patients} & \textbf{Eligible / Excluded} & \textbf{Pairs per patient (median / max)} \\
    \midrule
    Train & TREC 2021 & 9,302 & 60 & 4,348 / 4,954 & --- \\
    Development & TREC 2021 & 2,233 & 15 & 1,168 / 1,065 & --- \\
    Test & TREC 2022 & 6,975 & 50 & 3,939 / 3,036 & 116 / 398 \\
    \bottomrule
  \end{tabularx}
\end{table}

Trial text is built from the ClinicalTrials.gov XML snapshot used by the track, as sections: title, official title, conditions, interventions, keywords, summary, demographics and eligibility. The recipe reproduces all 26,162 distinct trial texts of the 2021 source file exactly. Inputs are short on the patient side (median 134 tokens, maximum 338) and long on the trial side (median 633, 95th percentile 1,603, maximum 3,578). The pointer model caps the patient state at 512 tokens and the combined state--branch input at 4,096; only 40 of the 18,510 pairs needed trimming, and the eligibility section is cut last. For label sanity, 54 ``eligible'' pairs whose patient age or sex contradicts the trial's own structured limits were removed from train and development; 82 such pairs in the test set are flagged but kept.

No test patient appears in train or development. Trial records can recur across years: 246 test pairs (3.53\%, 211 distinct trials) contain a trial also present in train or development. Their test label agrees with the training-majority label for that trial in 61.8\% of cases. This does not establish a simple label-memorization explanation, but the split is patient-disjoint rather than trial-disjoint. The backbones postdate the benchmark, so pretraining contamination remains possible.

The clinical initialization used at 0.8B is the released MedJev adapter and pointer head \cite{ref4}, trained for clinical-variable extraction from patient notes; it contains no TREC data. 

%======================================================================
\section{TrialJev and the comparison systems}\label{sec:4}

\subsection{TrialJev}\label{sub:4.1}

The patient note (the state) is tokenized and truncated from the head at 511 tokens. The question reads ``Is this patient eligible for the clinical trial below? Eligible means every inclusion criterion is met and no exclusion criterion applies.'', followed by the trial text and two options, ``Excluded: an inclusion criterion is not met or an exclusion criterion applies'' and ``Eligible: meets the inclusion criteria and no exclusion criterion applies''. Each question is rendered as the state followed by \texttt{[question] [option 1] </opt> [option 2] </opt> <decide>}, using rarely used Qwen special tokens as delimiters. A bilinear pointer head of dimension 256 compares the hidden state at \texttt{<decide>} with the hidden state at the end of each option span, and a softmax over the two scores gives the probabilities. Only the LoRA weights and the head are trained. On hybrid backbones the state is recomputed for each question during training. At serving time the state can be encoded once and each question reads its cache; we time this two-pass path (prefix, then branch) and a one-pass path that encodes the same sequence in a single pass, as in training.

Let $h_d$ denote the hidden state at the decision token and $h_i$ the hidden state at the end of option $i$. Two learned affine projections, $q$ and $k$, map these states to dimension $D = 256$. The head computes $s_i = k(h_i) \cdot q(h_d) / \sqrt{D}$. Normalizing across the options gives $P(i \mid p, t) = \operatorname{exp}(s_i) / \sum_j \operatorname{exp}(s_j)$; training minimizes $-\operatorname{log} P(y \mid p, t)$. The backbone is frozen apart from LoRA, and the pointer projections are trained. This separates the answer space from the vocabulary used for text generation.

We use Qwen3.5-0.8B-Base, Qwen3.5-9B-Base, Qwen3.5-27B and Qwen3.8-27B, all hybrid backbones with Gated DeltaNet layers. The first two are pretraining-only checkpoints; the two 27B backbones are post-trained checkpoints. At 0.8B we also initialize from the released MedJev adapter and head. ``Fresh'' refers to newly initialized task adapters and, for TrialJev, a new pointer head, not to an untrained backbone. Revisions and environments are in Appendices B and E.

\subsection{Training and checkpoint selection}\label{sub:4.2}

All models use LoRA of rank 16 with $\alpha$ = 32 and dropout 0.05 on the attention, MLP and Gated DeltaNet projection groups; AdamW with weight decay 0.01 and learning rate 1e-4 on a one-cycle schedule with 5\% warm-up; gradient clipping at 1.0; three epochs (1,746 optimizer steps) at effective batch 16. Checkpoints are saved periodically and selected by lowest development loss, not test accuracy. Additional test evaluations and post-hoc checks are disclosed in Appendix B.6. The seed is 0 unless stated; seeds 1 and 2 are added at 0.8B and 9B. Appendix C lists the settings and training cost.

\subsection{Comparison systems}\label{sub:4.3}

\textit{The A/B-logit control.} The prompt is the state, the question and trial, then \texttt{A. <first option>}, \texttt{B. <second option>} and \texttt{Answer:}. The probability is a softmax over the language-model head's logits for the tokens `` A'' and `` B'' at the last position, and training minimizes cross-entropy over those two logits, matching the pointer head's two-way loss. It adds no parameters beyond LoRA, uses the same data, recipe, record order and checkpoint selection, and is not conventional generative fine-tuning: nothing but the two label-token scores at one position is supervised. We also score it with the letters swapped (A = Eligible) as a diagnostic.

\textit{Prompted generation.} The un-adapted post-trained model receives its chat template with native thinking disabled and is asked to write ``Eligible'' or ``Excluded''. Greedy decoding allows at most eight new tokens. After case and boundary-punctuation normalization, any other output is invalid and counts as wrong. The 9B prompted model is the instruct checkpoint, whereas the adapted models use 9B-Base; this comparison does not isolate adaptation. At 27B the backbone checkpoints are the same. The 0.8B baseline uses Base with a plain prompt; no 0.8B chat condition was evaluated. A separate matched-prompt diagnostic scores the un-adapted backbone on the control's lettered prompt. Prompts are in Appendix D.2.

\textit{Fine-tuned LLM.} The fine-tuned LLM receives patient and trial text, word-valued options and an ``Answer:'' suffix without a chat template. The target is `` Eligible'' or `` Excluded'' followed by the end-of-sequence token. For an answer sequence $a$ of length $L$, its loss is the mean of $-\operatorname{log} P(a_k \mid \text{prompt}, a_{<k})$ over $k = 1,\ldots,L$, averaged across examples. Cross-entropy uses the full vocabulary, while prompt tokens are masked from the loss. This is conventional answer-only generative fine-tuning, not restricted A/B-logit training. We use the same data, adapter settings, optimizer schedule and development-loss selection recipe, but the prompt and objective differ from TrialJev. Greedy inference stops at the end token or eight new tokens. Accuracy measures the parsed written verdict; AUROC uses a separate two-way score from the first tokens of the two answer words. We train three seeds at 9B and one elsewhere. The 0.8B fine-tuned LLM starts fresh, so its initialization is not matched to the clinically initialized TrialJev.

\textit{Zero-shot explanation baseline.} The post-trained model (Qwen3.5-9B, Qwen3.5-27B or Qwen3.8-27B) is asked, with its chat template and thinking disabled, to reason briefly and end with exactly one line, ``Answer: Eligible'' or ``Answer: Excluded'' (Appendix D.3). Decoding is greedy with at most 512 new tokens, the verdict is the last such line, and output without one is invalid and counts as wrong. Because explanations are slow, this baseline is run on a 200-pair subset of the timing sample (4 pairs for each of the 50 patients), not on the full test set. It is a zero-shot baseline with a token cap, not a tuned or fine-tuned reasoner.

\textit{Hosted Jev.} TypeSafe's Jev (API-reported version jev-1.13.0) is queried through its API with the request our records already use: the patient note as the state and one typed yes/no question whose instructions contain the trial text and whose criteria are the two options; the label is not sent. The API returns a probability for ``true'' (eligible). We treat it as an external system used without adaptation by us; its size and training data are unknown. Latency is the end-to-end client-side time of one request, which is what a site calling the service experiences (Appendix B.10).

\subsection{Evaluation}\label{sub:4.4}

We report accuracy, macro-F1, AUROC, the Brier score (binary form) and expected calibration error (ECE, 10 bins). Because multiple patient–trial pairs share the same patient, we compute confidence intervals using patient-clustered bootstrapping, resampling the 50 test patients with replacement (10,000 repetitions for accuracy and 2,000 for hosted-Jev AUROC differences). Model comparisons use the same bootstrap samples to obtain paired confidence intervals. Reported p-values are two-sided. For multi-seed rows the metric is the seed mean, with the seed standard deviation given separately. Time is the median per pair at batch size 1 with GPU synchronization and no prefix reuse. For each model size all systems are timed on the same 992-pair sample (up to 20 pairs per patient) in one chain, one after another on one GPU, and the explanation baseline on its 200-pair subset; we report ratios to TrialJev's one-pass path within the chain, because absolute latency varied between runs of identical code (Appendix A.8). The pair count is stated for every result: 6,975 for full-test accuracy, 992 for timing, 200 for explanation. Intervals and p-values are not adjusted for multiple comparisons, and a non-significant difference is not evidence of equivalence. Because systems differ in their native operating points, we also report the score-returning system after the same development-fitted decision correction, in the form MedJev applied to Jev \cite{ref4}: a temperature (0.5 to 3.0) and a class-prior weight fitted on the TREC 2021 development split only, frozen, and applied once to the test scores.

%======================================================================
\section{Results}\label{sec:5}

\subsection{Main results across sizes and initializations}\label{sub:5.1}

\begin{table}[htbp]
  \centering
  \renewcommand{\thetable}{2}
  \caption{TREC 2022 test results (6,975 pairs, 50 patients). TrialJev uses the canonical two-pass evaluator; measured full-test accuracy changes by at most 0.09 points for the one-pass path at 9B--27B (Table A6). Brackets are 95\% patient-clustered bootstrap intervals. Three-seed rows give mean $\pm$ SD; the interval reflects patient sampling, not seed uncertainty. Other rows are seed 0. ``Control'' is the A/B-logit ablation. Bold marks the best value among the 9B--27B rows.}
  \label{tab:2}
  \footnotesize
  \setlength{\tabcolsep}{4pt}
  \begin{tabularx}{\linewidth}{P{1.32}P{1.31}P{1.89}P{0.79}P{0.69}P{0.51}P{0.49}}
    \toprule
    \textbf{Model} & \textbf{System} & \textbf{Accuracy [95\% CI]} & \textbf{Macro-F1} & \textbf{AUROC} & \textbf{Brier $\downarrow$} & \textbf{ECE $\downarrow$} \\
    \midrule
    Always ``eligible'' & --- & 56.47\% & --- & 0.500 & --- & --- \\
    0.8B, fresh & TrialJev (3 seeds) & 54.27 $\pm$ 0.95 [51.75, 57.11] & 51.2\% & 0.582 & 0.254 & 0.077 \\
    0.8B, fresh & control & 52.85 [47.91, 57.99] & 52.5\% & 0.567 & 0.252 & 0.070 \\
    0.8B, MedJev init & TrialJev (3 seeds) & 70.71 $\pm$ 2.64 [68.12, 73.36] & 70.4\% & 0.786 & 0.196 & 0.044 \\
    9B & TrialJev (3 seeds) & 78.27 $\pm$ 0.55 [75.80, 80.75] & 78.1\% & 0.862 & 0.158 & 0.063 \\
    9B & control (3 seeds) & 78.23 $\pm$ 2.74 [75.43, 80.92] & 77.9\% & 0.861 & 0.155 & \textbf{0.024} \\
    27B & TrialJev & 80.97 [78.14, 83.50] & 80.8\% & 0.885 & 0.145 & 0.089 \\
    27B & control & 81.25 [78.36, 83.81] & 81.0\% & 0.884 & \textbf{0.136} & 0.028 \\
    Qwen3.8-27B & TrialJev & \textbf{81.84} [79.04, 84.28] & \textbf{81.7\%} & \textbf{0.888} & 0.145 & 0.098 \\
    Qwen3.8-27B & control & 78.25 [74.89, 81.34] & 78.2\% & 0.887 & 0.146 & 0.028 \\
    \bottomrule
  \end{tabularx}
\end{table}

\begin{figure}[htbp]
  \centering
  \includegraphics[width=\linewidth]{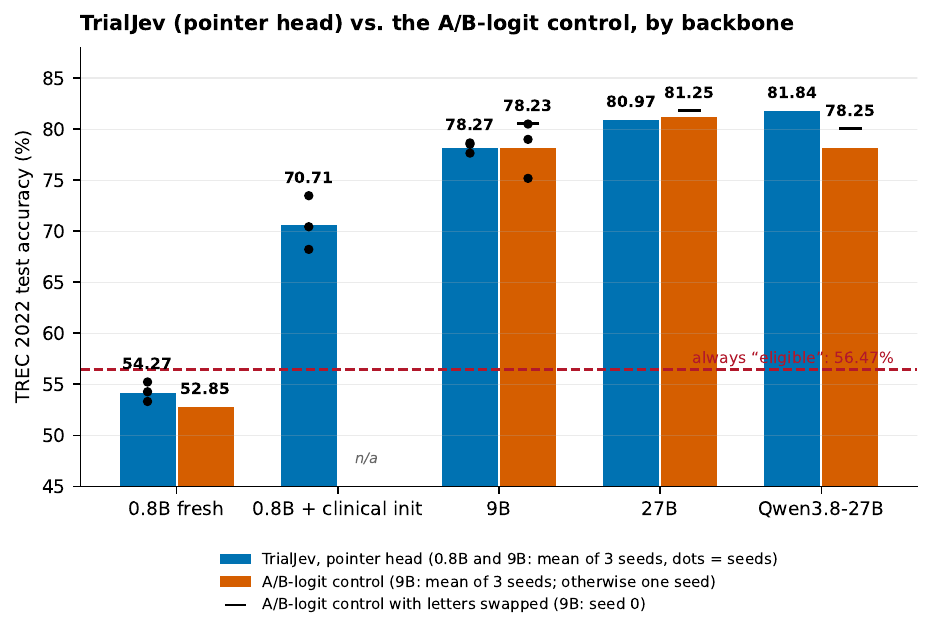}
  \renewcommand{\thefigure}{2}
  \caption{Test accuracy by backbone for TrialJev and the A/B-logit control (bars: mean of three seeds at 0.8B and, for both systems, at 9B, with seeds as dots; one seed otherwise). Short horizontal bars mark the control scored with the letters swapped (9B: seed 0). The dashed line is the always-``eligible'' accuracy.}
  \label{fig:2}
\end{figure}

TrialJev reaches 80.97\% on Qwen3.5-27B and 81.84\% on Qwen3.8-27B (AUROC 0.885 and 0.888), against a floor of 56.47\%. At 9B the three-seed mean is 78.27 $\pm$ 0.55\% (AUROC 0.862). With 50 test patients each interval spans about $\pm$3 points, so most neighboring rows of Table 2 cannot be separated; the comparisons below are paired. Accuracy rises from 9B to 27B by 2.70 points [+1.26, +4.25], against a single 27B seed. Qwen3.8-27B is 0.86 points above Qwen3.5-27B [+0.05, +1.61], p = 0.039, although on the 15-patient development set the order is reversed (77.88\% against 79.00\%), so we treat it as at most a small difference.

Initialization matters for the small pointer model. Fresh TrialJev at 0.8B averages 54.27\% accuracy with weak discrimination, whereas initialization from the released MedJev adapter and head raises the mean to 70.71\%: a gain of 16.44 points [+13.49, +19.31]. This result does not imply that clinical initialization is necessary for all small-model approaches. The fresh generative fine-tuned LLM reaches 74.05\% (Section 5.2). We did not separate the effects of transferring the clinical adapter and pointer head.

\subsection{Scoring versus writing a verdict}\label{sub:5.2}

The central comparison is with the conventional alternative: an LLM fine-tuned with the same data, adapter settings and checkpoint selection to write the verdict word (Section 4.3). Table 3 gives full-test accuracy and, from one timing chain per model size, latency on a 992-pair sample. Full-test AUROC and timing-sample accuracy are reported in Appendix A.15, Table A13.

\begin{table}[htbp]
  \centering
  \renewcommand{\thetable}{3}
  \caption{Screening accuracy and inference cost. Full-test accuracy uses 6,975 pairs and the canonical two-pass TrialJev evaluator. Mean $\pm$ SD denotes three seeds where available. The timing columns use seed-0 checkpoints on the same 992 pairs per backbone, batch 1, in one sequential chain on one GPU. Ratios are the fine-tuned LLM's median divided by TrialJev's one-pass median; full-test accuracy and subset latency are different measurements. The 0.8B initializations are unmatched (TrialJev: clinical; LLM: fresh). H100 NVL/bf16 is used at 9B--27B, RTX 6000 Ada/fp32 at 0.8B. Full serving-path details and subset accuracies are in Appendix A.15, Table A13; one-pass/two-pass prediction differences are in Table A6.}
  \label{tab:3}
  \footnotesize
  \begin{tabularx}{\linewidth}{P{0.79}P{0.92}P{1.19}P{1.15}P{0.95}}
    \toprule
    \textbf{Backbone} & \textbf{TrialJev accuracy} & \textbf{Fine-tuned LLM accuracy} & \textbf{TrialJev one-pass median ms} & \textbf{LLM / TrialJev latency} \\
    \midrule
    Qwen3.5-0.8B & 70.71 $\pm$ 2.64\% & 74.05\% & 54.0 & 1.85$\times$ \\
    Qwen3.5-9B & 78.27 $\pm$ 0.55\% & 78.34 $\pm$ 2.31\% & 106.2 & 2.45$\times$ \\
    Qwen3.5-27B & 80.97\% & 79.87\% & 177.6 & 2.62$\times$ \\
    Qwen3.8-27B & 81.84\% & 79.93\% & 178.2 & 2.70$\times$ \\
    \bottomrule
  \end{tabularx}
\end{table}

\begin{figure}[htbp]
  \centering
  \includegraphics[width=\linewidth]{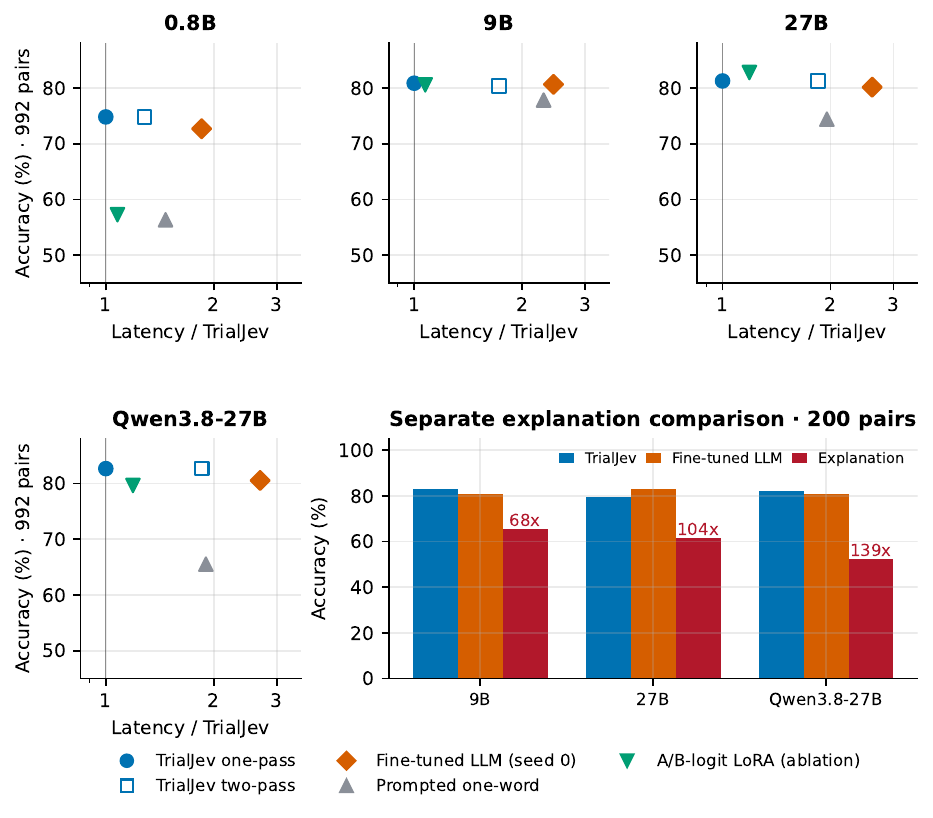}
  \renewcommand{\thefigure}{3}
  \caption{Four scatter panels compare accuracy and median latency on the 992-pair timing sample (seed-0 models), relative to TrialJev's one-pass path (logarithmic latency axis). The lower-right bar panel uses the separate 200-pair subset: all three systems' accuracies are measured on those same 200 pairs. Labels above the explanation bars give its median latency as a multiple of TrialJev's, not an accuracy ratio.}
  \label{fig:3}
\end{figure}

\textit{Accuracy against the fine-tuned LLM.} Across three 9B seeds, TrialJev and the fine-tuned LLM average 78.27\% and 78.34\%. At 27B their full-test accuracies are 80.97\% and 79.87\%, and on Qwen3.8-27B 81.84\% and 79.93\%. Seed-0 paired differences, TrialJev minus fine-tuned LLM, are $-$0.97 points [$-$4.56, +2.47], +1.10 [$-$0.57, +3.03] and +1.91 [+0.80, +3.06], respectively. Only the last unadjusted patient-bootstrap interval excludes zero. We describe similar observed accuracy, not demonstrated equivalence. Timing-sample paired comparisons and per-run fine-tuned LLM results are in Appendix A.11.

The fine-tuned LLM's first-token score has AUROC 0.860--0.864 at 9B, 0.879 at 27B and 0.888 on Qwen3.8-27B. This score measures ranking, not the written decision: at 9B the decoded verdict differs from thresholding it on 146--211 pairs per seed. Accuracy SD across three 9B seeds is 0.55 points for TrialJev and 2.31 for the fine-tuned LLM. The fine-tuned LLM's lower third-seed accuracy accompanies a lower predicted-eligible fraction, but we did not establish the cause; three seeds provide descriptive evidence, not a robustness guarantee.

\textit{Small models.} At 0.8B the fine-tuned LLM reaches 74.05\% on the full test set, against 73.48\% for the TrialJev seed that was timed and 70.71\% for the three-seed TrialJev mean. The fine-tuned LLM starts from a fresh adapter and TrialJev from the clinical initialization, so this is a seed-0 comparison and not a matched experiment. On the 992-pair sample TrialJev is 2.12 points above the fine-tuned LLM [$-$1.32, +5.84], and the fine-tuned LLM's median latency is 1.85 times TrialJev's. A fresh 0.8B fine-tuned LLM therefore learns the task where the fresh pointer and A/B-logit models stay at 52.9 to 54.3\% (Section 5.1); the fine-tuned LLM differs from them in its prompt, its objective and its answer tokens, and we did not isolate the cause. At this size TrialJev's case rests on speed and the clinical initialization, not on higher accuracy.

\textit{Time.} The fine-tuned LLM's median latency per pair was 2.45, 2.62 and 2.70 times TrialJev's one-pass path at 9B, 27B and Qwen3.8-27B, and 1.85 times at 0.8B; the prompted one-word model took 2.30, 1.96, 1.90 and 1.47 times. The one-pass path matters: TrialJev's two-pass path, used for the accuracy rows, was 1.28 to 1.85 times slower than its one-pass path, so against the two-pass path the fine-tuned LLM's median latency is 1.41 to 1.46 times. The control reading its A/B logits took 1.07 to 1.19 times the one-pass path. These ratios describe batch-1 inference in one implementation (Hugging Face generation with the model's own cache). The fine-tuned LLM emits three tokens including the end token, and its timed call also collects the output scores for its two-way score. The ratios are not engine-independent, they are not end-to-end workflow times, and batching was not evaluated. Because the A/B-logit control also scores without decoding and runs at nearly the same speed, the measured advantage belongs to single-pass scoring generally rather than to the pointer head specifically (Section 5.4).

\textit{Prompted and untrained models.} Table 4 gives the zero-shot alternatives: the same LLMs without fine-tuning, asked to write one word, and scored on lettered options without any adapter. At 9B the prompted model is the instruct checkpoint, not the Base model TrialJev starts from, so its result is not an adaptation effect.

\begin{table}[htbp]
  \centering
  \renewcommand{\thetable}{4}
  \caption{Zero-shot LLMs and MedJev as released (6,975 pairs; ratio to TrialJev's one-pass path in the same chain on the 992-pair sample, as in Table 3). Letter scoring has no timing because we did not time it. The 0.8B prompted model is the Base model with a plain prompt; no 0.8B chat condition was run.}
  \label{tab:4}
  \footnotesize
  \begin{tabularx}{\linewidth}{P{2.29}P{0.59}P{0.66}P{0.66}P{0.80}}
    \toprule
    \textbf{System} & \textbf{0.8B} & \textbf{9B} & \textbf{27B} & \textbf{Qwen3.8-27B} \\
    \midrule
    Always ``eligible'' & 56.47\% & 56.47\% & 56.47\% & 56.47\% \\
    Prompted LLM, one word: accuracy & 56.01\% & 77.35\% & 72.89\% & 61.66\% \\
    Prompted LLM, one word: median ms (ratio) & 79.2 (1.47) & 243.9 (2.30) & 347.1 (1.96) & 339.1 (1.90) \\
    Un-adapted, letter scoring & 55.70\% & 43.68\% & 64.00\% & not run \\
    MedJev as released, no TREC training & 56.59\% & -- & -- & -- \\
    TrialJev (Table 3) & 70.71\% & 78.27\% & 80.97\% & 81.84\% \\
    \bottomrule
  \end{tabularx}
\end{table}

Untrained models are at or near the majority floor at 0.8B, and the prompted LLM is 8.1 and 20.2 points below TrialJev at 27B and Qwen3.8-27B while taking 1.9 to 2.0 times as long. The 9B prompted instruct model (77.35\%) is close to TrialJev (78.27\%) but 2.3 times slower, and it is a different checkpoint. The released MedJev adapter and head, which were trained for clinical-variable extraction, predict ``eligible'' for 99.7\% of pairs (56.59\%, AUROC 0.562). Fine-tuning on TREC lifts that start to 70.71\%, while a fresh pointer head at 0.8B stays at 54.27\%, so at this size clinical initialization substantially improves the pointer model's observed screening accuracy under this recipe (Section 5.1).

\textit{Explain-then-answer.} We also timed a zero-shot explain-then-answer baseline on a 200-pair subset, with native thinking disabled and a 512-token cap (Appendix A.16). Its median latency was 68--139 times TrialJev's one-pass path and its accuracy 17.5--29.5 points lower on the same pairs, partly because 20 to 61 of the 200 outputs reached the cap without a verdict. Because the baseline is capped and untuned, we treat it as an illustration of the cost of per-pair explanation, not as an upper bound on reasoning-based matching.

\subsection{Comparison with a hosted decision model}\label{sub:5.3}

\textit{Hosted Jev under a common decision correction.} Hosted Jev (API-reported version jev-1.13.0), without adaptation by us, ranks pairs well (AUROC 0.882, versus 0.885 for TrialJev Qwen3.5-27B). Its native 0.5 cutoff predicts eligible for 23.7\% of pairs and yields 63.89\% accuracy. Applying the same development-only fitting procedure to hosted Jev and TrialJev gives Jev 81.28\%, TrialJev Qwen3.5-27B 80.99\% and TrialJev Qwen3.8-27B 81.88\% (Table 5). Their paired differences are $-$0.29 [$-$1.16, +0.62] and +0.60 [$-$0.30, +1.49] points: compatible with small differences, not proof of equivalence. TrialJev trails corrected Jev by 2.34 points at 9B and 10.47 at 0.8B. Its mean accuracy changes little under this correction (approximately 0.0--0.7 points), whereas hosted Jev gains 17.39 points. This does not validate TrialJev's probability calibration or a clinical cutoff. TrialJev has the highest observed corrected point accuracy among our tested score-based variants from 9B upwards, with only one seed at each 27B backbone; at 0.8B the fine-tuned LLM score is higher. The 15-patient development set makes the fitted correction uncertain. Figure 4 separates corrected score accuracy from the latency of the actual inference paths.

\begin{table}[htbp]
  \centering
  \renewcommand{\thetable}{5}
  \caption{Full-test accuracy under a common development-fitted decision correction (6,975 pairs). The same temperature/prior-mixing grid is fitted separately for each system and seed on 2,233 development pairs from 15 TREC 2021 patients, then frozen. Raw values use each system's native score decision rule; fine-tuned LLM values are first-token score decisions, not decoded-verdict accuracy. Corrected decisions consistently resolve an exact tie as excluded (Appendix B.8). Differences use unadjusted paired 95\% patient-bootstrap intervals, conditional on the fitted maps and observed seeds. Three-seed rows show mean $\pm$ sample SD; 27B rows have one seed. Fresh 0.8B classifiers are in Appendix A.1.}
  \label{tab:5}
  \footnotesize
  \begin{tabularx}{\linewidth}{P{1.11}P{0.69}P{1.04}P{1.16}}
    \toprule
    \textbf{System} & Raw native score decision & \textbf{Development-corrected} & \textbf{Corrected minus corrected Jev [95\% CI]} \\
    \midrule
    TrialJev 0.8B, MedJev init (3 seeds) & 70.71 $\pm$ 2.64 & 70.81 $\pm$ 2.32 & $-$10.47 [$-$13.22, $-$7.71] \\
    TrialJev 9B (3 seeds) & 78.27 $\pm$ 0.55 & 78.94 $\pm$ 1.21 & $-$2.34 [$-$3.60, $-$0.95] \\
    TrialJev Qwen3.5-27B & 80.97 & 80.99 & $-$0.29 [$-$1.16, +0.62] \\
    TrialJev Qwen3.8-27B & 81.84 & 81.88 & +0.60 [$-$0.30, +1.49] \\
    Fine-tuned LLM score 0.8B (fresh) & 74.05 & 72.89 & $-$8.39 [$-$10.86, $-$5.73] \\
    Fine-tuned LLM score 9B (3 seeds) & 78.41 $\pm$ 1.51 & 78.06 $\pm$ 0.65 & $-$3.22 [$-$4.51, $-$1.91] \\
    Fine-tuned LLM score Qwen3.5-27B & 80.34 & 80.34 & $-$0.93 [$-$2.67, +0.72] \\
    Fine-tuned LLM score Qwen3.8-27B & 80.62 & 80.62 & $-$0.66 [$-$1.73, +0.30] \\
    A/B-logit control 9B (3 seeds) & 78.23 $\pm$ 2.74 & 77.68 $\pm$ 2.23 & $-$3.60 [$-$4.97, $-$2.14] \\
    A/B-logit control Qwen3.5-27B & 81.25 & 79.54 & $-$1.73 [$-$3.26, $-$0.23] \\
    A/B-logit control Qwen3.8-27B & 78.25 & 79.23 & $-$2.05 [$-$3.42, $-$0.75] \\
    Hosted Jev (jev-1.13.0) & 63.89 & 81.28 & --- \\
    \bottomrule
  \end{tabularx}
\end{table}

\begin{figure}[htbp]
  \centering
  \includegraphics[width=\linewidth]{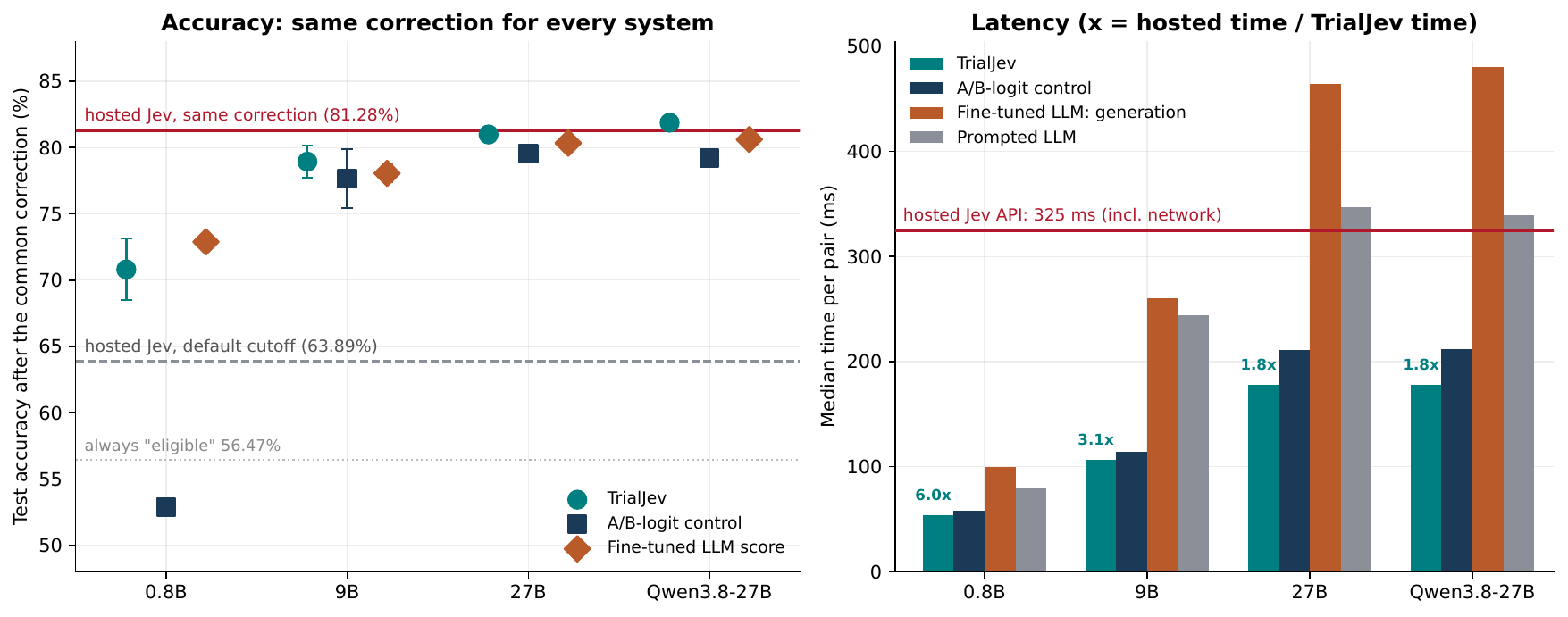}
  \renewcommand{\thefigure}{4}
  \caption{Corrected score accuracy and local inference latency are distinct measurements. Left: full-test accuracy after the common development-fitted correction; bars show training-seed SD where three seeds exist, not patient confidence intervals. Hosted Jev reference lines show corrected and raw accuracy. Fine-tuned LLM points use first-token scores, not written verdicts. Right: median local inference latency on the same 992 pairs, batch 1, using seed-0 models. Fine-tuned LLM timing measures verdict generation, not score-only classification. All local timings exclude tokenization and model loading. Hardware is RTX 6000 Ada/fp32 at 0.8B and H100 NVL/bf16 at 9B--27B; across-size bars are not hardware-controlled comparisons. The red line is hosted Jev's median request time (325 ms), which includes the network and service hardware we do not control; the labels above the TrialJev bars give the hosted request time divided by TrialJev's local time (6.0x at 0.8B, 3.1x at 9B, 1.8x at both 27B models).}
  \label{fig:4}
\end{figure}

\textit{Hosted request latency and local inference time.} One-worker hosted Jev calls on the 992-pair sample took 325 ms at the median (p90 360 ms; p99 455 ms), including the network and service execution on unknown hardware. TrialJev's local one-pass inference medians were approximately 178 ms at 27B, 106 ms at 9B and 54 ms at 0.8B, excluding tokenization and model loading. TrialJev's local model time is therefore 1.8 times lower than the hosted median at both 27B backbones, 3.1 times lower at 9B and 6.0 times lower at 0.8B. The two figures have different timing boundaries, so they are measured request times and not a hardware-controlled comparison; the speed gap does not depend on the accuracy correction, which only changes the cutoff applied to returned probabilities. Of 2,638 pairs accidentally sent to the API twice, 1,384 returned a different rounded probability and 23 changed decision at the raw 0.5 cutoff. This is an observational repeat-call diagnostic, not a controlled determinism comparison with TrialJev. Local weights can be pinned; repeated same-path local determinism was not measured (Appendix A.13).

\subsection{Readout ablation: what the pointer head adds}\label{sub:5.4}

The A/B-logit control attains similar observed accuracy on Qwen3.5: 78.23\% versus TrialJev's 78.27\% at 9B over three seeds, and 81.25\% versus 80.97\% at 27B. Paired accuracy intervals include zero, as do the AUROC differences (Appendix A.2). These results show that a pointer head is not uniquely required to obtain strong binary screening performance. They do not establish statistical equivalence or a head-only causal effect, because prompt serialization also differs.

On Qwen3.8-27B, TrialJev is 3.58 points more accurate [+1.97, +5.32], despite similar AUROC (0.888 versus 0.887). The control predicts eligible less often than the gold prevalence. This is consistent with an operating-point difference, not proof of its cause. A post-hoc step-400 check does not remove the gap (Appendix A.6); one seed per configuration prevents attribution to the backbone alone.

\textit{Calibration.} The control has lower ECE: 0.024 versus 0.063 at 9B, 0.028 versus 0.089 at 27B, and 0.028 versus 0.098 on Qwen3.8-27B. Paired intervals for these differences exclude zero. Figure 5 shows predominantly under-confident TrialJev predictions at 9B and 27B, with over-confidence in the top 27B bin (confidence 0.98, accuracy 0.94). We did not inspect Qwen3.8 reliability bin by bin. A normalized score is not automatically a calibrated clinical probability; a deployment threshold and any calibration map must be fitted using development data.

\begin{figure}[htbp]
  \centering
  \includegraphics[width=\linewidth]{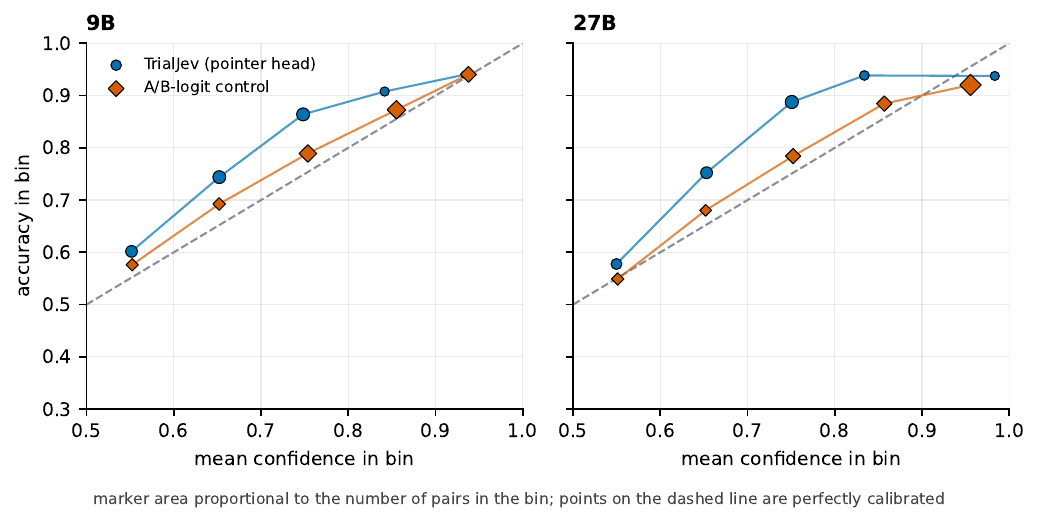}
  \renewcommand{\thefigure}{5}
  \caption{Reliability of TrialJev and the A/B-logit control at 9B and 27B (seed-0 models; 10 equal-width confidence bins; marker area proportional to the number of pairs).}
  \label{fig:5}
\end{figure}

Swapping the control's letter assignments changes 2--5\% of predictions. TrialJev avoids assigning labels to vocabulary letters, but option-order sensitivity remains untested. Its potential to handle richer option sets and multiple questions is architectural capacity, not an efficiency advantage demonstrated by this binary benchmark.

\textit{Where the accuracy comes from.} Taken together, the results separate three ingredients. At 27B the backbone already supplies much of the ranking signal: the un-adapted Qwen3.5-27B reaches AUROC 0.860 on the control's lettered prompt, against 0.884 after adaptation. Adaptation there is accompanied by a large shift in the operating point: accuracy rises from 64.00\% to 81.25\% (+17.25 points [+10.77, +24.41]), and the un-adapted model predicts eligible for only 24.4\% of pairs against 56.5\% in the gold labels (Appendix A.7). The readout matters little for accuracy on Qwen3.5 at 9B and 27B, where TrialJev, the A/B-logit control and the fine-tuned LLM lie within 1.4 points of each other, but it determines latency and calibration. The picture changes with weaker starting points: the pretraining-only 9B-Base model ranks poorly without adaptation (AUROC 0.559), and at 0.8B the objective and initialization decide whether the task is learned at all (fresh TrialJev 54.27\%, fresh fine-tuned LLM 74.05\%, MedJev-initialized TrialJev 70.71\%). These are observational contrasts across configurations, not a controlled decomposition.

%======================================================================
%======================================================================
\section{Using TrialJev for first-pass screening}\label{sec:6}

TrialJev's intended role is between retrieval and review. A retriever proposes candidate trials \cite{ref11}; TrialJev assigns eligibility scores; a clinician or downstream explainer examines shortlisted and uncertain candidates. The scores already order candidates usefully: within each test patient's judged pool, 9.4 of the top 10 trials are eligible for both 27B TrialJev models (precision@10 0.936--0.940), against 5.9 for random order (Appendix A.10). That pool is easier than a retriever's output, since it contains only already-relevant trials, 59\% of them eligible; performance on a noisier candidate pool must be established before deployment. The measured latency advantage makes this staged design plausible, but we did not evaluate the complete workflow or its effect on staff time.

A useful first pass should preserve potentially eligible trials, not merely maximize accuracy. A development-fitted decision correction was evaluated here, but no recall-targeted threshold or reviewer-workload evaluation was performed. Future operating points should be selected on development patients for a prespecified recall target, then assessed using eligible-trial recall, shortlist size and reviewer workload on
held-out patients. A low score must not be treated as authority to exclude a patient from clinical review.

\textit{Deployment.} TrialJev can run locally using openly available backbone weights; task heads are released on Hugging Face (see Availability). The tested configurations use one H100 NVL at 9B--27B or an RTX 6000 Ada at 0.8B. Inference need not send patient inputs to an external model API and avoids a vendor per-request fee, although local hardware, maintenance and governance costs remain. The clinically initialized
0.8B model offers a smaller footprint: it averages 70.71\% and is faster than its fine-tuned LLM comparator on the same card, at the cost of about ten accuracy points relative to 27B. Local processing does not remove privacy review, security controls or data-use obligations, none of which was evaluated here.

The principal trade-off is explainability. TrialJev does not identify a failed criterion, cite evidence or provide the decision traces offered by generative and formal systems \cite{ref1,ref13}. Its score is a
prioritization signal, not a substitute for those outputs, and its use on real clinical records has not been evaluated.

\textit{Next steps.} The pointer readout supports structured decisions beyond the binary task studied here. A third option, not relevant, could extend it to reranking a retrieved candidate pool, provided appropriate labels and evaluation are available. Criterion-level typed questions over a shared patient state could expose inclusion and exclusion decisions without free-text decoding, and encoding a long patient
record once for many such questions could add a further efficiency gain; on these short TREC notes that saving is small (Appendix A.8). These extensions are untested and require independent evidence for accuracy, explainability and efficiency.

%======================================================================
\section{Limitations}\label{sec:7}

\textbf{Generalization.} Evaluation uses one benchmark with 50 synthetic test patients and excludes irrelevant trials. Test patients are disjoint from training, but trial records can recur, and backbone pretraining contamination is unmeasured. A new patient-held-out collection, full-retrieval candidates and real-record evaluation are needed to establish external validity.

\textbf{Training and uncertainty.} Patient-bootstrap intervals do not include training-seed variation. There are three seeds at 9B but only one per 27B configuration and one for the 0.8B fine-tuned LLM. Development-loss selection uses 15 patients and no hyperparameter search. Base versus instruct comparisons and the unmatched clinical initialization at 0.8B limit causal interpretation. The pointer, A/B control and fine-tuned LLM also use different input serialization and objectives. Post-hoc checks are labeled separately rather than used to replace selected results.

\textbf{Generation baseline scope.} The fine-tuned LLM is conventionally fine-tuned to produce a verdict, not a rationale. The explanation baseline is zero-shot, with native thinking disabled and a 512-token cap. These experiments are not upper bounds on generative matching or reasoning. Explanation quality, longer budgets, few-shot prompts and a fine-tuned reasoner were not tested. The hosted Jev comparison uses one model version (jev-1.13.0) of an API whose size, training data and serving hardware are unknown; the decision correction for all systems was fitted on only 15 development patients; one call per pair was used although repeated calls differ slightly; and hosted latency is end-to-end, including the network.

\textbf{Latency and safety.} Timing uses batch 1, shared nodes and one sequential chain per size, with three cluster chains running concurrently on separate GPUs. Fine-tuned LLM timing includes Hugging Face generation and score collection; an optimized engine or batching may change the ratios. Absolute latency varied between runs for an unidentified reason. Tokenization and model loading are outside the per-pair timing window, so the figures are not end-to-end workflow costs. One-pass and two-pass scores are numerically close but not identical. TrialJev requires calibration and operating-point validation and produces no supporting evidence. It is a research model, not a clinically validated enrollment or exclusion system.

%======================================================================
\section{Conclusion}\label{sec:8}

TrialJev provides a local, single-pass patient--trial eligibility score without decoding text. On the binary TREC 2022 evaluation, its two 27B configurations achieve 80.97--81.84\% accuracy. A conventionally fine-tuned LLM has similar observed screening accuracy, while its median inference latency is 2.45--2.70 times TrialJev's under matched-pair, batch-1 evaluation at 9B--27B. The speed result is specific to these inference implementations. A common development-fitted decision correction places both 27B TrialJev models near hosted Jev in observed accuracy, with paired intervals that do not establish equivalence; smaller TrialJev models trail it. The A/B-logit control shows that scoring rather than writing accounts for much of the efficiency, while a pointer head offers a structured interface whose richer uses remain to be tested. Clinical initialization helps the 0.8B pointer recipe, but fresh conventional generative fine-tuning achieves higher observed accuracy at that size. These findings motivate a staged workflow with a local scorer and human or explanatory-model review, not automatic enrollment or exclusion. Recall-targeted evaluation, real clinical data and independent retrieval and criterion-level benchmarks are the next tests of practical value. Task-trained weights are released alongside the documented configurations.

\section*{Availability}

The project repository is \url{https://github.com/oibrahi1/TrialJev}. It contains the code to rebuild the dataset, train and evaluate every model, and recompute the reported statistics. Processed TREC data and per-pair judgments are not redistributed; the dataset must be rebuilt from the original TREC and ClinicalTrials.gov sources. TrialJev is released as an open-weight model family on Hugging Face, together with the fine-tuned LLM and A/B-logit comparison models, in three repositories: \url{https://huggingface.co/omareng/TrialJev}, \url{https://huggingface.co/omareng/TrialJev-FineTunedLLM} and \url{https://huggingface.co/omareng/TrialJev-ABLogit}. The release covers all 23 task-trained model configurations and reported training seeds in this study, including fresh and MedJev-initialized 0.8B TrialJev. Task weights are distributed as LoRA adapters and, for TrialJev, the corresponding pointer heads; users load the upstream backbone separately. Hosted Jev is a third-party API baseline and is not part of our weight release. Code and adapters are released under Apache-2.0.

\textbf{Affiliation.} TrialJev is independent research that adopts the design of the open-weight MedJev and kev releases (Apache-2.0). It is not affiliated with or endorsed by TypeSafe AI, the developer of Jev.

\section*{Ethics and data statement}

Training and evaluation used synthetic TREC patient descriptions and public trial records. The clinically initialized 0.8B model also starts from the released MedJev weights, which were trained on public case-report-derived text. No institutional patient records were used. The model is a research artifact and must not be used for clinical decisions or enrollment.

\noindent \dag{} Vendor description; the performance and speed figures are the vendor's claims and were not verified. All other entries were checked against the cited source (abstract, repository page, or the track overview).

\clearpage

%======================================================================
\appendix

\section{Extended results}\label{app:A}

This appendix keeps the full evidence behind the main text. Here ``pointer head'' and ``pointer'' refer to TrialJev, and ``A/B-logit LoRA'', ``LM-logit classifier'' and ``control'' to the A/B-logit control of Section 4.3. Tables and sections carry the numbering of this appendix.

\subsection{Full headline table and seed spread}\label{sub:A.1}

\begin{table}[htbp]
  \centering
  \renewcommand{\thetable}{A1}
  \caption{TREC 2022 test results (6,975 pairs, 50 patients). ``A/B-logit LoRA'' is the LM-logit classifier of Appendix B.4: a classifier whose loss uses only the two label-token logits, not conventional generative fine-tuning. Brackets are 95\% patient-clustered bootstrap intervals. Rows marked ``3 seeds'' give the mean $\pm$ sd of three seeds, with the interval for the seed mean; per-seed values are in Appendix A.9. Other rows are one seed (seed 0). Bold marks the best value in each column among the 9B to 27B rows, excluding the letter-swapped diagnostics.}
  \label{tab:A1}
  \footnotesize
  \setlength{\tabcolsep}{4pt}
  \begin{tabularx}{\linewidth}{P{1.65}P{1.33}P{1.62}P{0.79}P{0.69}P{0.46}P{0.46}}
    \toprule
    \textbf{Model} & \textbf{Readout} & \textbf{Accuracy [95\% CI]} & \textbf{Macro-F1} & \textbf{AUROC} & \textbf{Brier $\downarrow$} & \textbf{ECE $\downarrow$} \\
    \midrule
    Always ``eligible'' & --- & 56.47\% & --- & 0.500 & --- & --- \\
    0.8B, fresh & pointer (3 seeds) & 54.27 $\pm$ 0.95 [51.75, 57.11] & 51.2\% & 0.582 & 0.254 & 0.077 \\
    0.8B, fresh & A/B-logit LoRA & 52.85 [47.91, 57.99] & 52.5\% & 0.567 & 0.252 & 0.070 \\
    0.8B, MedJev init & pointer (3 seeds) & 70.71 $\pm$ 2.64 [68.12, 73.36] & 70.4\% & 0.786 & 0.196 & 0.044 \\
    9B & pointer (3 seeds) & 78.27 $\pm$ 0.55 [75.80, 80.75] & 78.1\% & 0.862 & 0.158 & 0.063 \\
    9B & A/B-logit LoRA (3 seeds) & 78.23 $\pm$ 2.74 [75.43, 80.92] & 77.9\% & 0.861 & 0.155 & \textbf{0.024} \\
    27B & pointer & 80.97 [78.14, 83.50] & 80.8\% & 0.885 & 0.145 & 0.089 \\
    27B & A/B-logit LoRA & 81.25 [78.36, 83.81] & 81.0\% & 0.884 & \textbf{0.136} & 0.028 \\
    Qwen3.8-27B & pointer & \textbf{81.84} [79.04, 84.28] & \textbf{81.7\%} & \textbf{0.888} & 0.145 & 0.098 \\
    Qwen3.8-27B & A/B-logit LoRA & 78.25 [74.89, 81.34] & 78.2\% & 0.887 & 0.146 & 0.028 \\
    \textit{9B, letters swapped (seed 0)} & \textit{A/B-logit LoRA} & \textit{80.54 [77.65, 83.31]} & \textit{80.2\%} & \textit{0.868} & \textit{0.148} & \textit{0.056} \\
    \textit{27B, letters swapped} & \textit{A/B-logit LoRA} & \textit{81.82 [78.87, 84.44]} & \textit{81.5\%} & \textit{0.885} & \textit{0.134} & \textit{0.022} \\
    \textit{Qwen3.8-27B, letters swapped} & \textit{A/B-logit LoRA} & \textit{80.06 [76.90, 82.92]} & \textit{80.0\%} & \textit{0.888} & \textit{0.142} & \textit{0.022} \\
    \bottomrule
  \end{tabularx}

  \smallskip
  \raggedright\footnotesize The AUROC of the always-``eligible'' predictor is 0.5 by definition.
\end{table}

Each interval spans about $\pm$3 points. With 50 test patients, most differences between neighboring rows of Table A1 are smaller than the uncertainty of either row, so the comparisons below use paired intervals. Seeds matter at 9B: the three LM-logit seeds score 80.50, 79.00 and 75.18\%, a range of 5.3 points, and the three pointer-head seeds 78.52, 77.65 and 78.65\%, a range of 1.0 point. The 27B and Qwen3.8-27B rows are single seeds, so differences of a few points between them cannot be separated from seed variation.

\subsection{Paired comparisons}\label{sub:A.2}

\begin{table}[htbp]
  \centering
  \renewcommand{\thetable}{A2}
  \caption{Paired differences (A $-$ B) with 95\% patient-clustered intervals; p is two-sided and shown as <0.001 when it is below that value. ``Swapped'' rows use the letter-swapped LM-logit pass, a diagnostic. Where a side is marked ``3 seeds'', the value is the seed mean and the interval reflects patient sampling only.}
  \label{tab:A2}
  \footnotesize
  \begin{tabularx}{\linewidth}{P{1.59}P{1.33}P{0.38}P{0.85}P{0.85}}
    \toprule
    \textbf{Comparison (A $-$ B)} & \textbf{Accuracy, points [95\% CI]} & \textbf{p} & \textbf{ECE [95\% CI]} & \textbf{AUROC [95\% CI]} \\
    \midrule
    9B, 3 seeds: A/B-logit LoRA $-$ pointer & $-$0.05 [$-$1.34, +1.16] & 0.95 & $-$0.039 [$-$0.045, $-$0.019] & $-$0.001 [$-$0.009, +0.006] \\
    27B: A/B-logit LoRA $-$ pointer & +0.27 [$-$0.35, +0.87] & 0.39 & $-$0.061 [$-$0.077, $-$0.036] & $-$0.001 [$-$0.005, +0.002] \\
    Qwen3.8-27B: A/B-logit LoRA $-$ pointer & $-$3.58 [$-$5.32, $-$1.97] & <0.001 & $-$0.070 [$-$0.083, $-$0.040] & $-$0.001 [$-$0.008, +0.005] \\
    9B, seed 0 only: A/B-logit LoRA $-$ pointer & +1.98 [$-$0.45, +4.43] & 0.12 & $-$0.055 [$-$0.074, $-$0.027] & +0.007 [$-$0.002, +0.015] \\
    9B, seed 0: A/B-logit LoRA (swapped) $-$ pointer & +2.02 [+0.17, +3.92] & 0.031 & $-$0.022 [$-$0.040, $-$0.002] & +0.008 [$-$0.001, +0.016] \\
    27B: A/B-logit LoRA (swapped) $-$ pointer & +0.85 [+0.09, +1.61] & 0.029 & $-$0.068 [$-$0.084, $-$0.030] & +0.000 [$-$0.004, +0.004] \\
    Qwen3.8-27B: A/B-logit LoRA (swapped) $-$ pointer & $-$1.78 [$-$3.00, $-$0.67] & 0.001 & $-$0.076 [$-$0.088, $-$0.047] & +0.000 [$-$0.007, +0.007] \\
    pointer: 27B $-$ 9B (3 seeds) & +2.70 [+1.26, +4.25] & <0.001 & +0.027 [+0.006, +0.044] & +0.023 [+0.013, +0.034] \\
    A/B-logit LoRA: 27B $-$ 9B (3 seeds) & +3.02 [+1.80, +4.12] & <0.001 & +0.004 [$-$0.011, +0.013] & +0.023 [+0.013, +0.033] \\
    pointer: Qwen3.8-27B $-$ Qwen3.5-27B & +0.86 [+0.05, +1.61] & 0.039 & +0.008 [$-$0.001, +0.017] & +0.003 [$-$0.003, +0.009] \\
    A/B-logit LoRA: Qwen3.8-27B $-$ Qwen3.5-27B & $-$3.00 [$-$4.80, $-$1.32] & <0.001 & $-$0.000 [$-$0.013, +0.015] & +0.003 [$-$0.006, +0.011] \\
    27B A/B-logit LoRA $-$ Qwen3.8-27B pointer & $-$0.59 [$-$1.44, +0.27] & 0.20 & $-$0.070 [$-$0.087, $-$0.041] & $-$0.004 [$-$0.012, +0.003] \\
    0.8B: MedJev init $-$ fresh (pointer, 3 seeds) & +16.44 [+13.49, +19.31] & <0.001 & $-$0.033 [$-$0.069, +0.005] & +0.205 [+0.163, +0.247] \\
    0.8B fresh: A/B-logit LoRA $-$ pointer (3 seeds) & $-$1.42 [$-$4.16, +1.22] & 0.30 & $-$0.007 [$-$0.028, +0.011] & $-$0.015 [$-$0.033, +0.004] \\
    \bottomrule
  \end{tabularx}
\end{table}

On Qwen3.5 the paired accuracy differences are not distinguishable from zero in these runs. At 9B the three-seed means are 78.23\% and 78.27\% (difference $-$0.05 points, interval [$-$1.34, +1.16]), and at 27B the LM-logit classifier is ahead by 0.27 points with an interval that includes zero. The seed-0 pair alone shows +1.98 points at 9B, but that gap sits inside the LM-logit classifier's own seed spread, which is why seed means are the headline. The letter-swapped diagnostics at 9B (seed 0) and 27B put the LM-logit classifier slightly higher (+2.02 and +0.85, intervals just excluding zero); we treat them as diagnostics because they use the unselected letter order.

On Qwen3.8-27B the result differs. The pointer head is 3.58 points more accurate than the LM-logit classifier [+1.97, +5.32], and 1.78 points more accurate [+0.67, +3.00] than the letter-swapped A/B-logit pass. The two models have similar observed AUROC (0.888 and 0.887). The accuracy gap is consistent with a different operating point at the 0.5 threshold; it does not establish the cause or identical pair ordering. In the normal letter order the LM-logit model recovers 0.712 of eligible pairs and 0.874 of excluded ones and predicts 45.7\% of pairs eligible against 56.5\% in the gold labels, whereas the pointer head is balanced (0.811 and 0.828). The A/B-logit Qwen3.8 checkpoint was the earliest one saved, step 200 (development loss 0.5130 against 0.517 at step 400), but scoring its step-400 checkpoint afterwards, as a post-hoc check, gives 78.11\% (80.29\% letter-swapped), a change of $-$0.14 points [$-$1.11, +0.91] in the normal order, and the pointer head stays ahead by 3.73 points [+1.96, +5.56] (Appendix A.6). The A/B-logit classifier on Qwen3.8-27B scores 3.00 points below the same classifier on Qwen3.5-27B [$-$4.80, $-$1.32], whereas the pointer head scores 0.86 points above its Qwen3.5 counterpart. With one seed per cell we cannot say whether this reflects the backbone or seed variation of the size seen at 9B; scoring step 400 did not remove the gap, but other checkpoints and selection rules were not tested.

In the 9B, 27B and Qwen3.8-27B comparisons, no AUROC difference exceeds 0.008 in magnitude and every interval includes zero. The observed Qwen3.5 accuracies are similar, but these findings do not demonstrate equivalence or non-inferiority. On Qwen3.8-27B the pointer head is ahead at the default threshold.

Calibration differs in the same direction in every comparison. ECE is lower for the LM-logit classifier by 0.039 at 9B (three seeds), 0.061 at 27B and 0.070 at Qwen3.8-27B, and all three intervals exclude zero. The LM-logit classifier's ECE is stable across its five runs (0.022 to 0.028 in the normal order), while the pointer head's ranges from 0.051 to 0.098. The reliability diagram (Figure 5, seed-0 models) shows the reason: the pointer head is mostly under-confident (at 9B accuracy is above confidence in every bin; at 27B in every bin except the top one, where confidence 0.98 meets accuracy 0.94), while the LM-logit classifier stays close to the diagonal. Similar aggregate AUROC does not imply identical rankings. The observed calibration differences concern confidence scaling; we did not isolate their cause.

We did not test why the LM-logit classifier is better calibrated. One explanation is that a fine-tuned language model already represents the two answers in its output space, so an extra learned head adds parameters and optimization difficulty without adding information. The pointer head is built for settings with many questions, a long shared state and arbitrary option texts, which this task does not use (Section 6).

\subsection{Effect of model size}\label{sub:A.3}

With three seeds at 9B, accuracy rises from 9B to 27B by 2.70 points for the pointer head [+1.26, +4.25] and by 3.02 points for the LM-logit classifier [+1.80, +4.12], and AUROC rises by 0.023 for both; the 27B values are single seeds. The two 27B backbones are not consistently ordered: Qwen3.8-27B is 0.86 points above Qwen3.5-27B for the pointer head ([+0.05, +1.61], p = 0.039, no detectable AUROC difference) and 3.00 points below it for the LM-logit classifier (Appendix A.2). On the 15-patient development set the pointer-head order reverses (accuracy 77.88\% vs 79.00\%, loss 0.4918 vs 0.4871), each is a single seed, and the test effect barely excludes zero. We therefore treat Qwen3.8-27B as at most slightly better for the pointer head and do not rank the two backbones. The 27B LM-logit model cannot be told apart in accuracy from the Qwen3.8-27B pointer-head model ($-$0.59, [$-$1.44, +0.27]) and is much better calibrated (ECE 0.028 vs 0.098). The 27B backbones are post-trained while the 0.8B and 9B ones are Base, and the 27B cells have one seed, so Table A1 compares the listed systems and is not a clean parameter-count experiment.

\subsection{Small models and clinical initialization}\label{sub:A.4}

At 0.8B, fresh fine-tuning does not exceed the majority-class accuracy with either classifier and discriminates weakly. The three pointer-head seeds reach 54.27, 55.23 and 53.32\% (mean 54.27 [51.75, 57.11]), all below the 56.47\% floor, with AUROC 0.58, which is weak discrimination. The A/B-logit LoRA classifier reaches 52.85\% (AUROC 0.57). Starting the pointer-head model from the released MedJev adapter and head gives 73.48, 68.22 and 70.44\% (mean 70.71 [68.12, 73.36], AUROC 0.79). The paired gain is 16.44 points [13.49, 19.31], large next to the seed standard deviations of 1.0 and 2.6, and the weakest clinically initialized seed is far above the best fresh seed. We report the three-seed mean rather than selecting the unusually strong first seed (73.48\%). The two settings also differ in AUROC by 0.205 [0.163, 0.247].

The gain cannot be attributed to the adapter or the head, because the LM-logit classifier has no matching start. Loading the released adapter into the A/B-logit model would separate the two; we did not run that experiment.

\begin{figure}[htbp]
  \centering
  \includegraphics[width=\linewidth]{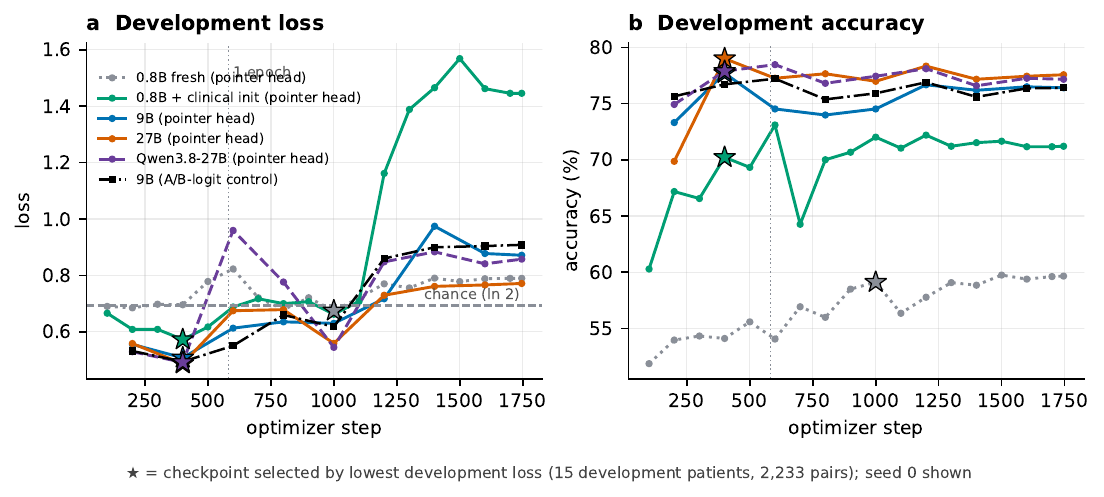}
  \renewcommand{\thefigure}{A1}
  \caption{Development loss (a) and accuracy (b) for the pointer-head runs (seed 0) and the 9B A/B-logit LoRA classifier. Stars mark the checkpoints selected by lowest development loss.}
  \label{fig:A1}
\end{figure}

\subsection{Operating point, variation across patients, and ensembles}\label{sub:A.5}

At the 0.5 threshold, the 9B LM-logit classifier (seed 0) leans toward ``eligible''. It recovers 0.874 of eligible pairs and 0.715 of excluded ones, and predicts 61.8\% of pairs eligible against 56.5\% in the gold labels. The 9B pointer head (seed 0) is balanced (0.773 and 0.801). At 27B both readouts are balanced (A/B-logit: 0.814 and 0.810; pointer: 0.813 and 0.805). On Qwen3.8-27B the LM-logit classifier leans the other way, toward ``excluded'' (0.712 and 0.874; 45.7\% predicted eligible), and the pointer head is balanced (0.811 and 0.828). The lean of the LM-logit classifier therefore changes from run to run, and a deployment would set the threshold on development data for its own recall target, which we did not do. Swapping which option is called A leaves 95.4\% (9B, seed 0), 98.0\% (27B) and 94.9\% (Qwen3.8-27B) of predictions unchanged.

Accuracy varies a good deal across patients. The per-patient median is 0.81 to 0.86 across the six models of Appendix A.2 (seed 0 at 9B) and the interquartile range about 0.74 to 0.88. The worst patient scores 0.30 to 0.44 and the best 0.93 to 1.00, with a standard deviation of 0.10 to 0.12. This spread, together with only 50 patients, explains why the intervals are wide.

Ensembles were evaluated as a post-hoc check on the five seed-0 systems that existed when the analysis was fixed (9B and 27B A/B-logit LoRA, 9B, 27B and Qwen3.8-27B pointer). Averaging the predicted probabilities of fixed model sets did not beat the best single model. The two LM-logit models (9B and 27B) reach 81.82 [78.79, 84.60] with ECE 0.031, the three pointer-head models 81.33 [78.45, 83.99] with ECE 0.100, and all five 81.84 [78.74, 84.64] with ECE 0.079. Ensembling gives no free gain here, and it carries over the pointer head's under-confidence.

\subsection{Checkpoint selection and post-hoc step-400 scoring}\label{sub:A.6}

Development loss was lowest at step 400 (about 0.7 epochs) in seven of the ten runs at 9B and above: five of the six seed-0 runs and both readouts' seed-1 runs at 9B. It was lowest at step 200 for the 9B LM-logit seed 2 and the Qwen3.8-27B LM-logit run, and at step 1000 for the 9B pointer-head seed 2. In the step-400 runs it rose afterwards while development accuracy stayed similar, which fits growing overconfidence or overfitting (Figure A1). The development loss fluctuates a good deal from checkpoint to checkpoint (for pointer-head seed 2 at 9B between 0.54 and 1.04), and with 15 development patients its minimum is a noisy choice of checkpoint. The two LM-logit runs sent to step 200 are the two lowest-scoring LM-logit runs at 9B and above (75.18\% and 78.25\%), so we scored their step-400 checkpoints on the test set afterwards, as a post-hoc check that was not used to select anything. Test accuracy barely moves: 9B seed 2 gives 75.48\% at step 400 against 75.18\% at step 200 (+0.30 points [$-$2.30, +3.09]), and Qwen3.8-27B 78.11\% against 78.25\% ($-$0.14 [$-$1.11, +0.91]), with 80.29\% against 80.06\% in the swapped order. Calibration is worse at step 400 for the 9B run (ECE 0.079 against 0.026), as the loss rule would favor. Scoring step 400 therefore did not change the picture for these two runs; other checkpoints and selection rules were not tested.

For the fresh 0.8B runs the rule behaves differently. The selected checkpoints fall at steps 100 to 1000, where development loss is 0.67 to 0.72, essentially the chance value (ln 2 = 0.693), while development accuracy climbs to 65 to 67\% later in training in some runs. The fresh-0.8B result therefore depends on the selection rule. We kept one pre-specified rule and did not score later checkpoints on the test set. A development-accuracy rule is possible as a labeled post-hoc analysis.

\begin{table}[htbp]
  \centering
  \renewcommand{\thetable}{A3}
  \caption{Post-hoc step-400 scoring of the two LM-logit runs whose development-loss minimum was at step 200 (TREC 2022 test; scored after the main analysis and not used for selection). The ``selected'' rows are the headline entries.}
  \label{tab:A3}
  \footnotesize
  \begin{tabularx}{\linewidth}{P{2.02}P{1.27}P{1.08}P{0.71}P{0.46}P{0.46}}
    \toprule
    \textbf{Run} & \textbf{Checkpoint} & \textbf{Accuracy [95\% CI]} & \textbf{Macro-F1} & \textbf{Brier} & \textbf{ECE} \\
    \midrule
    9B A/B-logit LoRA, seed 2 & step 200 (selected) & 75.18\% [71.36, 78.56] & 75.17\% & 0.1715 & 0.0262 \\
    9B A/B-logit LoRA, seed 2 & step 400 (post hoc) & 75.48\% [72.44, 78.34] & 75.48\% & 0.1855 & 0.0793 \\
    Qwen3.8-27B A/B-logit LoRA & step 200 (selected) & 78.25\% [74.89, 81.34] & 78.24\% & 0.1462 & 0.0280 \\
    Qwen3.8-27B A/B-logit LoRA & step 400 (post hoc) & 78.11\% [74.96, 81.07] & 78.09\% & 0.1492 & 0.0388 \\
    Qwen3.8-27B A/B-logit LoRA, letters swapped & step 200 (selected) & 80.06\% [76.90, 82.92] & 79.96\% & 0.1425 & 0.0220 \\
    Qwen3.8-27B A/B-logit LoRA, letters swapped & step 400 (post hoc) & 80.29\% [77.29, 82.97] & 80.13\% & 0.1392 & 0.0248 \\
    \bottomrule
  \end{tabularx}
\end{table}

\subsection{Prompted generation, matched-prompt zero-shot and the 0.8B label token}\label{sub:A.7}

\begin{table}[htbp]
  \centering
  \renewcommand{\thetable}{A4}
  \caption{Generation and zero-shot baselines on the full test set (6,975 pairs). Strict scoring: an answer that is not exactly one of the two permitted words or letters is invalid and counts as wrong. The last column is the same condition with the two options in the other order (A = Eligible for the letter prompt). The matched-prompt rows have no adapter and score the A/B logits of the fine-tuning prompt. Brackets are 95\% patient-clustered intervals.}
  \label{tab:A4}
  \footnotesize
  \begin{tabularx}{\linewidth}{P{1.06}P{1.52}P{0.82}P{0.71}P{0.98}P{0.91}}
    \toprule
    \textbf{Condition} & \textbf{Model} & \textbf{Accuracy [95\% CI]} & \textbf{Macro-F1} & \textbf{Invalid answers} & \textbf{Other order: accuracy [95\% CI]} \\
    \midrule
    Zero-shot chat, writes the word & Qwen3.5-9B (instruct) & 77.35\% [74.41, 80.31] & 77.3\% & 0 & 74.71\% [70.92, 78.58] \\
    Zero-shot chat, writes the word & Qwen3.5-27B & 72.89\% [66.77, 77.83] & 72.8\% & 0 & 71.54\% [65.24, 76.66] \\
    Zero-shot chat, writes the word & Qwen3.8-27B & 61.66\% [54.78, 68.37] & 59.6\% & 1 & 58.61\% [51.56, 65.60] \\
    Zero-shot plain, writes the word & Qwen3.5-9B-Base & 44.24\% [37.31, 51.35] & 39.9\% & 1,149 (16.5\%) & not run \\
    Zero-shot letters, matched prompt, scored & Qwen3.5-0.8B-Base & 55.70\% [48.26, 62.94] & 38.5\% & 0 & 46.72\% [44.05, 49.49] \\
    Zero-shot letters, matched prompt, scored & Qwen3.5-9B-Base & 43.68\% [35.89, 51.73] & 30.7\% & 0 & 56.89\% [48.93, 64.71] \\
    Zero-shot letters, matched prompt, scored & Qwen3.5-27B & 64.00\% [57.05, 70.34] & 62.6\% & 0 & 61.52\% [54.46, 68.20] \\
    A/B-logit LoRA, reads the logits & Qwen3.5-0.8B-Base (fresh, seed 0) & 52.85\% [47.91, 57.99] & 52.5\% & 0 & not run \\
    A/B-logit LoRA, emits its label token & Qwen3.5-9B-Base (seed 0) & 80.50\% [77.11, 83.63] & 79.8\% & 0 & 80.54\% [77.65, 83.31] \\
    A/B-logit LoRA, emits its label token & Qwen3.5-27B & 81.25\% [78.36, 83.81] & 81.0\% & 0 & 81.82\% [78.87, 84.44] \\
    \bottomrule
  \end{tabularx}
\end{table}

The A/B-logit LoRA classifier that reads the two letter logits in one pass makes exactly the same predictions as the same classifier allowed to emit its label token: of 6,975 pairs, none differs at 9B or at 27B, and the unrestricted next-token choice was always `` A'' or `` B''. The 27B A/B-logit LoRA row of Table A1 and the seed-0 9B A/B-logit LoRA row of Table A7 are therefore the same as the fine-tuned rows above. This holds only where the adapter has learned the task. The fresh 0.8B LM-logit adapter, whose development loss was at chance, does not reliably emit a label token: on the 992-pair timing sample its greedy output was an empty string after special tokens were removed for 534 pairs (53.8\%), which we read as the model preferring a token other than A or B, such as end of text, although the script does not record which token it was; strict accuracy fell to 23.59\% against 57.26\% when the same adapter's A/B logits were scored. The 458 pairs where it did write a letter agree with the scored answer in all cases. At 9B and 27B the unrestricted choice was always one of the two letters. Re-scoring the 9B and 27B checkpoints on the full test set reproduced the earlier saved predictions pair by pair.

The matched-prompt rows separate the effect of adaptation from that of the prompt. On the identical lettered prompt and backbone, adding the fine-tuned adapter raises 27B accuracy by 17.25 points [+10.77, +24.41] in the normal order and 20.30 points [+13.28, +27.98] in the swapped order. At 9B, where the backbone is the same Base checkpoint, the gap is 36.82 points [+27.84, +45.76] and 23.66 points [+15.68, +32.23]. The un-adapted 9B-Base model and the 0.8B model are dominated by letter bias: the 9B model predicts ``eligible'' for 0.3\% of pairs in one order and 99.5\% in the other, the 0.8B model for 96.4\% and 45.7\%, and their accuracy follows the share of the class that the favored letter happens to name. The 27B model is less biased (24.4\% and 21.6\% predicted eligible) but still below the zero-shot word-writing condition (64.00\% against 72.89\%, a difference of 8.89 points [+5.80, +12.36]). The 8.4-point gap between the adapted 27B control and prompted word writing is therefore the conservative comparison, since word writing is the stronger of the two zero-shot formats. At 0.8B, fresh adaptation gives no benefit over zero-shot scoring (52.85\% against 55.70\%, difference $-$2.85 points [$-$14.61, +9.21]), which fits the weak discrimination reported in Appendix A.4.

Ranking and accuracy tell different stories at 27B. The un-fine-tuned 27B model already ranks pairs well: AUROC 0.860 [0.830, 0.885], against 0.884 [0.858, 0.907] after fine-tuning (difference +0.024 [+0.006, +0.043]). A large part of its accuracy deficit on this prompt therefore comes from where the 0.5 threshold falls, that is, from its letter bias (24.4\% of pairs predicted eligible against 56.5\% in the gold labels), and only a smaller part from worse ranking. We did not tune a threshold for the un-fine-tuned model, so we cannot say what accuracy it would reach with one. At 9B-Base and 0.8B the un-fine-tuned model ranks poorly (AUROC 0.559 and 0.505), and fine-tuning adds 0.308 [+0.275, +0.344] at 9B.

Writing a verdict zero-shot is clearly weaker than the fine-tuned classifiers, and depends on the model and option order. At 27B the prompted LLM and A/B-logit control share a backbone checkpoint. The control emitting its one label token scores 8.36 points higher [+3.87, +14.40] in the normal order and 10.28 points higher [+5.39, +16.76] in the swapped order. This gap combines the effect of adaptation with that of the prompt and answer format, which differ between the two (Appendix B.7). At 9B the gaps are 3.15 [+0.38, +5.79] and 5.84 points [+2.20, +9.40], but the zero-shot model is the instruct checkpoint and the control starts from Base, so they are not attributable to adaptation. The zero-shot models also change their answers with the option order: accuracy falls by 1.35 points [0.83, 1.85] at 27B, 2.64 [0.76, 4.64] at 9B and 3.05 [2.13, 4.21] on Qwen3.8-27B when the order is swapped, with 94 to 97\% of predictions unchanged, whereas the adapted 27B control differs by 0.57 points between the orders (higher in the swapped order; interval [0.15, 1.04]) with 98.0\% of predictions unchanged. Zero-shot quality on Qwen3.8-27B is 11.23 points below Qwen3.5-27B [+7.18, +16.27] in the normal order, although the Qwen3.8 pointer head is the most accurate fine-tuned model; the zero-shot ordering of backbones does not carry over to the adapted one. The Base model prompted without the chat template does not follow the answer format in 16.5\% of cases and is below the majority floor.

For context, the released MedJev adapter and head with no TREC training score 56.59\% (AUROC 0.562) and predict ``eligible'' for 99.7\% of pairs. An earlier, cruder probe that read the A/B letter logits of the un-fine-tuned models (Appendix A.9, Table A8) agrees with the matched-prompt rows: 43.66\% against 43.68\% for Qwen3.5-9B-Base and 63.90\% against 64.00\% for Qwen3.5-27B (the probe used the chat template for post-trained models), and gave 68.66\% and 57.63\% for the Qwen3.5-9B and Qwen3.8-27B chat models.

\subsection{Timing details}\label{sub:A.8}

\textit{Note.} Tables 3 and 4 in Section 5.2 report the final same-chain timing comparisons. This subsection retains earlier, separate timing runs for the serving-path analysis and zero-shot comparison. Absolute latencies differed between runs, so latency ratios are interpreted only within the same timing run.

\begin{table}[htbp]
  \centering
  \renewcommand{\thetable}{A5}
  \caption{Median time per pair on the 992-pair timing sample (batch 1), with the ratio to the A/B-logit LoRA classifier reading the logits in the same run. ``Emits its label token'' is that same classifier allowed to output its selected label token, not conventional generative fine-tuning. 9B and 27B: one run per model size on one H100 NVL GPU, bf16, conditions executed one after another. 0.8B: a separate run on (RTX 6000 Ada, fp32); its pointer head is the MedJev-initialized checkpoint (seed 0) and its A/B-logit model the fresh adapter, so only the ratios, not the milliseconds, may be compared with the other columns. Models are the seed-0 fine-tuned models; the zero-shot 9B model is the instruct checkpoint. ``New tokens'' is the mean number of generated tokens.}
  \label{tab:A5}
  \footnotesize
  \setlength{\tabcolsep}{4pt}
  \begin{tabularx}{\linewidth}{P{1.93}P{1.66}P{1.57}P{0.62}P{0.53}P{0.54}P{0.57}P{0.58}}
    \toprule
    \textbf{Condition} & \textbf{New tokens} & \textbf{0.8B ms} & \textbf{0.8B ratio} & \textbf{9B ms} & \textbf{9B ratio} & \textbf{27B ms} & \textbf{27B ratio} \\
    \midrule
    A/B-logit LoRA, reads the logits (one pass) & 0 & 62.4 & 1.00 & 114.9 & 1.00 & 209.0 & 1.00 \\
    A/B-logit LoRA, emits its label token & 1 & 62.8 (54\% empty) & 1.01 & 113.6 & 0.99 & 209.7 & 1.00 \\
    Pointer head, two-pass path (A) & --- & 66.2 & 1.06 & 176.7 & 1.54 & 312.3 & 1.49 \\
    Pointer head, one-pass path (B) & --- & 51.3 & 0.82 & 104.0 & 0.90 & 172.0 & 0.82 \\
    Zero-shot chat, writes the word & 5.5 (9B), 3.4 (27B) & --- & --- & 244.4 & 2.13 & 340.1 & 1.63 \\
    Zero-shot plain (Base), writes the word & 3.8 & --- & --- & 181.5 & 1.58 & --- & --- \\
    \bottomrule
  \end{tabularx}
\end{table}

Paths A and B of the pointer head were timed on the same pairs, interleaved in one process. Path A is the canonical full-test evaluator: it encodes the patient prefix, then runs a branch pass from a copy of the cached state. Path B encodes the same sequence in one pass, as in training. At 0.8B the two paths gave the same prediction on all 992 pairs. The other rows are separate runs, executed one after another on the same GPU. No zero-shot chat condition was evaluated at 0.8B in this earlier experiment.

Three results follow. First, letting the A/B-logit classifier emit its selected label token instead of reading the logits costs nothing measurable (0.99 to 1.00 of the scoring time), because the answer is a single token and no prediction changes (Appendix A.7). Second, prompting the same backbone without the task adapter to write a verdict takes 1.63 times as long at 27B and 2.13 times at 9B, with 3.4 and 5.5 generated tokens on average. In the separate full-test-set run, where the conditions shared the node with other jobs, the same comparison on the same GPU gave 1.49 (27B: 299.9 against 201.0 ms) and 2.51 (9B: 195.9 against 77.9 ms). We measured only short verdicts of a few tokens. The explanation baseline of Appendix A.16 (Table A14b) was timed separately.

Third, the pointer head's speed depends on the serving path. The two-pass path used for the accuracy results took 1.49 to 1.54 times as long as reading the A/B logits at 9B and 27B, and 1.06 times as long at 0.8B, where the model is small enough that per-call overhead dominates. The one-pass path took 0.82 to 0.90 times as long, which makes the pointer head about as fast as reading the A/B logits and possibly 10 to 18\% faster; the 0.8B run shows the same pattern (one-pass path 0.82, two-pass path 1.06). That comparison rests on separate runs on the same GPU and not on interleaved timing, so we do not claim more than parity. The one-pass path is not identical to the two-pass path. Over all 6,975 pairs it changes 12 to 20 predictions (0.17 to 0.29\%) and moves accuracy by at most 0.09 points (Table A6), while the two-pass path reproduces the previously saved predictions exactly.

\begin{table}[htbp]
  \centering
  \renewcommand{\thetable}{A6}
  \caption{Pointer-head serving paths on the full test set (6,975 pairs), timed on the same pairs in the same process. Path A reproduces the earlier saved predictions and probabilities exactly (0 differences).}
  \label{tab:A6}
  \footnotesize
  \setlength{\tabcolsep}{4pt}
  \begin{tabularx}{\linewidth}{P{1.61}P{0.97}P{0.97}P{1.04}P{1.08}P{0.73}P{0.73}P{0.87}}
    \toprule
    \textbf{Backbone} & \textbf{Accuracy, path A} & \textbf{Accuracy, path B} & \textbf{Pairs with different prediction} & \textbf{Largest probability difference} & \textbf{Median ms, A} & \textbf{Median ms, B} & \textbf{A / B (median of pairwise ratios)} \\
    \midrule
    Qwen3.5-9B-Base & 78.52\% & 78.49\% & 20 (0.29\%) & 0.023 & 177.1 & 64.9 & 2.69 \\
    Qwen3.5-27B & 80.97\% & 81.00\% & 12 (0.17\%) & 0.045 & 327.8 & 170.6 & 1.93 \\
    Qwen3.8-27B & 81.84\% & 81.75\% & 12 (0.17\%) & 0.020 & 328.7 & 167.9 & 1.95 \\
    \bottomrule
  \end{tabularx}
\end{table}

Absolute latency was not stable between runs, which is why the paper compares conditions only within a run. The 9B A/B-logit classifier took 114.9 ms median on the 992 pairs in the timing run and 77.8 ms on the same pairs in the full-test-set run, with identical code and weights. A quiet-node calibration gave 111.6, 113.9, 114.0 and 113.0 ms on the four GPUs, so no single GPU explains the difference. At 9B the two-pass pointer path was steady (176.7 and 177.2 ms on the same pairs in the two runs) while the scoring paths changed by 32\% (A/B-logit classifier) and 38\% (one-pass pointer path, 104.0 and 64.7 ms), so run-to-run variation did not affect all paths equally. We do not know the cause. The absolute values in Table A5 describe one run each and should not be set against numbers from other hardware or load conditions.

Each one-pass classifier decision is one forward pass of known input length, without decoding or verdict parsing. The fine-tuned LLM is a separate generative system. In the earlier zero-shot runs, the chat model returned an invalid answer for 1 of 6,975 pairs at Qwen3.8-27B and none at 9B and 27B; parsing failures were rare for the chat template and common (16.5\%) for the Base prompt (Appendix A.7).

Training wall time (Appendix C) was lower for A/B-logit LoRA than for the pointer head by 13 to 20\% in matched runs (9B seeds 0 to 2: 2.77 to 3.01 h against 3.33 to 3.77 h; 27B: 8.1 h against 9.3 h), on shared machines.

The pointer-head design encodes the patient once and branches per trial from the cached prefix, but the saving is small on this data because patients are short: a mean of 136 tokens against 749 for the trial branch. Relative to re-encoding the patient for every trial, caching processes 1.16 times fewer tokens at N = 10 and at most 1.18 times fewer for N $\geq$ 100 (S + N$\cdot$B versus N$\cdot$(S + B)). It would matter for long patient records such as full EHR histories. At TREC note lengths we make no speed claim from it. The within-backbone latency comparison concerns scoring versus decoding. Cross-size and cross-hardware efficiency were not controlled.

At the small end, the 0.8B model with clinical initialization reaches a mean of 70.7\%. On the separate 0.8B timing run (RTX 6000 Ada, fp32) its one-pass path took 51.3 ms per pair and its two-pass path 66.2 ms, against 62.4 ms for the 0.8B A/B-logit classifier in the same run (Table A5). The 27B models reach 81 to 82\% on an H100 NVL. The small model gives up about ten accuracy points in exchange for modest hardware, which may suit a fast first-pass screen.

\subsection{Per-run test results and zero-shot probes}\label{sub:A.9}

\begin{figure}[htbp]
  \centering
  \includegraphics[width=\linewidth]{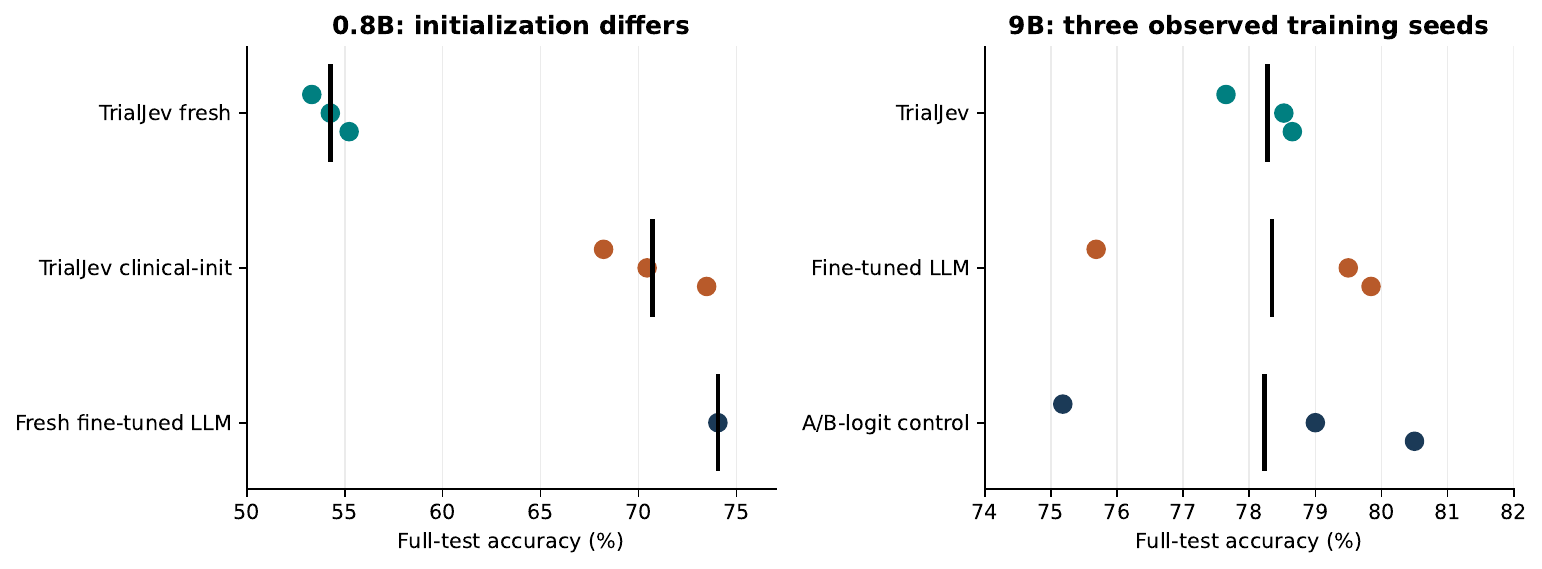}
  \renewcommand{\thefigure}{A2}
  \caption{Observed test accuracy across training seeds, on all 6,975 pairs. Each point is one selected checkpoint; the horizontal segment is the mean. The 0.8B panel contrasts fresh and clinically initialized TrialJev with a fresh fine-tuned LLM; initialization is not matched. The 9B panel includes three seeds for TrialJev, the fine-tuned LLM and the A/B-logit control. Seed spreads are descriptive, not a test of robustness or variance differences. Exact per-run values remain in Tables A7 and A10.}
  \label{fig:A2}
\end{figure}

\begin{table}[htbp]
  \centering
  \renewcommand{\thetable}{A7}
  \caption{All scored runs (TREC 2022 test). Brier is the conventional binary form (half the evaluator's two-class sum).}
  \label{tab:A7}
  \footnotesize
  \begin{tabularx}{\linewidth}{P{3.36}P{0.33}P{0.68}P{0.71}P{0.46}P{0.46}}
    \toprule
    \textbf{Run} & \textbf{Step} & \textbf{Accuracy} & \textbf{Macro-F1} & \textbf{Brier} & \textbf{ECE} \\
    \midrule
    0.8B pointer, fresh, seed 0 & 1000 & 54.27\% & 53.87\% & 0.2511 & 0.0881 \\
    0.8B pointer, fresh, seed 1 & 100 & 55.23\% & 47.77\% & 0.2454 & 0.0151 \\
    0.8B pointer, fresh, seed 2 & 500 & 53.32\% & 51.88\% & 0.2662 & 0.1276 \\
    0.8B pointer, MedJev init, seed 0 & 400 & 73.48\% & 72.97\% & 0.1897 & 0.0852 \\
    0.8B pointer, MedJev init, seed 1 & 500 & 68.22\% & 68.21\% & 0.2065 & 0.0340 \\
    0.8B pointer, MedJev init, seed 2 & 500 & 70.44\% & 70.13\% & 0.1903 & 0.0124 \\
    0.8B A/B-logit LoRA, seed 0 & 200 & 52.85\% & 52.50\% & 0.2517 & 0.0703 \\
    9B pointer, seed 0 & 400 & 78.52\% & 78.36\% & 0.1611 & 0.0778 \\
    9B pointer, seed 1 & 400 & 77.65\% & 77.34\% & 0.1595 & 0.0586 \\
    9B pointer, seed 2 & 1000 & 78.65\% & 78.59\% & 0.1527 & 0.0512 \\
    9B A/B-logit LoRA, seed 0 & 400 & 80.50\% & 79.83\% & 0.1431 & 0.0228 \\
    9B A/B-logit LoRA, seed 1 & 400 & 79.00\% & 78.70\% & 0.1489 & 0.0224 \\
    9B A/B-logit LoRA, seed 2 & 200 & 75.18\% & 75.17\% & 0.1715 & 0.0262 \\
    9B A/B-logit LoRA, seed 0, letters swapped & 400 & 80.54\% & 80.17\% & 0.1479 & 0.0562 \\
    27B pointer & 400 & 80.97\% & 80.75\% & 0.1450 & 0.0893 \\
    27B A/B-logit LoRA & 400 & 81.25\% & 81.03\% & 0.1362 & 0.0282 \\
    27B A/B-logit LoRA, letters swapped & 400 & 81.82\% & 81.54\% & 0.1338 & 0.0218 \\
    Qwen3.8-27B pointer & 400 & 81.84\% & 81.66\% & 0.1452 & 0.0977 \\
    Qwen3.8-27B A/B-logit LoRA & 200 & 78.25\% & 78.24\% & 0.1462 & 0.0280 \\
    Qwen3.8-27B A/B-logit LoRA, letters swapped & 200 & 80.06\% & 79.96\% & 0.1425 & 0.0220 \\
    \bottomrule
  \end{tabularx}
\end{table}

\begin{table}[htbp]
  \centering
  \renewcommand{\thetable}{A8}
  \caption{Zero-shot letter-logit probes (no adapter or head; the probability of the letter ``A'' or ``B'' read at the answer position; one option order; TREC 2022 test, 6,975 pairs). Base models get a plain prompt and post-trained models the chat template. These probes are cruder than the generation baselines of Table A4.}
  \label{tab:A8}
  \footnotesize
  \begin{tabular}{lll}
    \toprule
    \textbf{Model} & \textbf{Prompt} & \textbf{Accuracy} \\
    \midrule
    Qwen3.5-9B-Base & plain & 43.66\% \\
    Qwen3.5-9B (instruct) & chat & 68.66\% \\
    Qwen3.5-27B & chat & 63.90\% \\
    Qwen3.8-27B & chat & 57.63\% \\
    \bottomrule
  \end{tabular}
\end{table}

\subsection{Ranking within the judged-relevant pool}\label{sub:A.10}

\textit{Exploratory ranking inside the judged pool.} Without ranking-specific training, both systems put eligible trials near the top of a patient's judged candidate pool: precision@10 is 0.936 to 0.944 for the 27B-scale systems and 0.921 to 0.924 at 9B (metrics computed per seed, then averaged), against 0.590 for random order; nDCG@10 is 0.93 to 0.95. TrialJev and the control differ by at most 0.013 in precision@10 and nDCG@10, with intervals that include zero; the one exception is mean reciprocal rank on Qwen3.8-27B, where TrialJev is 0.04 lower [$-$0.08, $-$0.01]. With 59\% eligible trials on average across patients, this restricted pool has a high random baseline; near-ceiling top-k results here should not be extrapolated to noisier retrieved candidates. This analysis reranks the judged-relevant pool, from which irrelevant trials have been removed; it does not evaluate retrieval over the full collection (Table A9).

For each of the 50 test patients, the judged-relevant trials (eligible and excluded; irrelevant trials are absent from this binary dataset) are ranked by P(eligible), with eligible as the relevant class, and the metrics are averaged over patients (mean pool 140 trials, 59\% eligible). This is a post-hoc reranking analysis without ranking-specific training, not retrieval over the full collection. For multi-seed systems, metrics are computed separately for each seed and then averaged; this avoids turning the seeds into an unreported ensemble. At 9B the seed standard deviation is 0.008 (TrialJev) and 0.012 (control) for precision@10, and 0.008 and 0.010 for nDCG@10. Intervals resample patients. The random-order row is the expected value over 200 shuffles per patient.

\begin{table}[htbp]
  \centering
  \renewcommand{\thetable}{A9}
  \caption{Ranking within the judged-relevant pool (precision@10, nDCG@10 with binary gains, mean reciprocal rank of the first eligible trial, average precision).}
  \label{tab:A9}
  \footnotesize
  \begin{tabularx}{\linewidth}{P{1.47}P{0.86}P{0.95}P{0.86}P{0.86}}
    \toprule
    \textbf{System} & \textbf{P@10} & \textbf{nDCG@10} & \textbf{MRR} & \textbf{MAP} \\
    \midrule
    TrialJev 9B (3 seeds) & 0.921 [0.887, 0.950] & 0.930 [0.899, 0.957] & 0.980 [0.957, 0.997] & 0.882 [0.847, 0.912] \\
    Control 9B (3 seeds) & 0.924 [0.893, 0.951] & 0.934 [0.905, 0.959] & 0.983 [0.963, 0.997] & 0.884 [0.850, 0.914] \\
    TrialJev 27B & 0.940 [0.908, 0.966] & 0.945 [0.914, 0.970] & 0.970 [0.930, 1.000] & 0.894 [0.857, 0.926] \\
    Control 27B & 0.936 [0.904, 0.964] & 0.945 [0.919, 0.969] & 0.990 [0.970, 1.000] & 0.897 [0.864, 0.926] \\
    TrialJev Qwen3.8-27B & 0.936 [0.902, 0.964] & 0.941 [0.909, 0.968] & 0.960 [0.920, 0.990] & 0.897 [0.862, 0.928] \\
    Control Qwen3.8-27B & 0.944 [0.910, 0.972] & 0.954 [0.926, 0.978] & 1.000 [1.000, 1.000] & 0.901 [0.870, 0.930] \\
    Zero-shot letters 27B (no adapter) & 0.916 [0.872, 0.952] & 0.927 [0.891, 0.958] & 0.980 [0.950, 1.000] & 0.872 [0.832, 0.907] \\
    Random order (expected) & 0.590 [0.527, 0.651] & 0.589 [0.526, 0.650] & 0.735 [0.686, 0.783] & 0.609 [0.548, 0.668] \\
    \bottomrule
  \end{tabularx}
\end{table}

Paired differences (A $-$ B, patient bootstrap): TrialJev 9B (3 seeds) - Control 9B (3 seeds): P@10 $-$0.003 [$-$0.017, +0.009], nDCG@10 $-$0.004 [$-$0.016, +0.008], MRR $-$0.003 [$-$0.017, +0.007], MAP $-$0.003 [$-$0.009, +0.003]; TrialJev 27B - Control 27B: P@10 +0.004 [$-$0.008, +0.016], nDCG@10 $-$0.000 [$-$0.017, +0.015], MRR $-$0.020 [$-$0.060, +0.020], MAP $-$0.003 [$-$0.011, +0.003]; TrialJev Qwen3.8-27B - Control Qwen3.8-27B: P@10 $-$0.008 [$-$0.024, +0.008], nDCG@10 $-$0.013 [$-$0.030, +0.002], MRR $-$0.040 [$-$0.080, $-$0.010], MAP $-$0.004 [$-$0.014, +0.003]; Control 27B - Zero-shot letters 27B (no adapter): P@10 +0.020 [$-$0.000, +0.044], nDCG@10 +0.018 [$-$0.002, +0.041], MRR +0.010 [$-$0.020, +0.050], MAP +0.025 [+0.013, +0.039]. The 59\% eligible prevalence creates a high random baseline and limited headroom, so these metrics separate learned scores from random order more clearly than they separate TrialJev from the control.

\subsection{Fine-tuned LLM: per-run results}\label{sub:A.11}

\begin{table}[htbp]
  \centering
  \renewcommand{\thetable}{A10}
  \caption{Fine-tuned LLM on the full test set (6,975 pairs; strict scoring; no invalid answers in any run). AUROC is computed from the first-step scores of the two answer words; ``written differs from score'' counts pairs where the written verdict differs from thresholding that score at 0.5.}
  \label{tab:A10}
  \footnotesize
  \setlength{\tabcolsep}{4pt}
  \begin{tabularx}{\linewidth}{P{1.88}P{0.89}P{0.92}P{0.77}P{0.91}P{0.92}P{0.89}P{0.82}}
    \toprule
    \textbf{Run} & \textbf{Accuracy} & \textbf{Macro-F1} & \textbf{Eligible recall} & \textbf{Excluded recall} & \textbf{Predicted eligible} & \textbf{AUROC (first token)} & \textbf{Written differs from score} \\
    \midrule
    0.8B, fresh, seed 0 & 74.05\% & 73.0\% & 83.0\% & 62.4\% & 63.3\% & 0.802 & 0 \\
    9B, seed 0 & 79.50\% & 78.5\% & 89.7\% & 66.3\% & 65.3\% & 0.864 & 146 \\
    9B, seed 1 & 79.84\% & 79.0\% & 88.4\% & 68.7\% & 63.5\% & 0.862 & 149 \\
    9B, seed 2 & 75.68\% & 75.7\% & 67.0\% & 86.9\% & 43.6\% & 0.860 & 211 \\
    27B, seed 0 & 79.87\% & 79.8\% & 76.7\% & 84.0\% & 50.3\% & 0.879 & 139 \\
    Qwen3.8-27B, seed 0 & 79.93\% & 79.9\% & 75.7\% & 85.4\% & 49.1\% & 0.888 & 144 \\
    \bottomrule
  \end{tabularx}

  \smallskip
  \raggedright\footnotesize The gold labels are 56.5\% eligible. The 9B seed mean is 78.34\% with SD 2.31. All numbers are computed from the saved per-pair predictions.
\end{table}

Paired timing-sample differences (TrialJev one-pass minus fine-tuned LLM; 992 pairs) are +0.20 [$-$2.43, +2.84] at 9B, +1.11 [$-$1.01, +3.33] at 27B and +2.12 [+0.60, +3.78] on Qwen3.8-27B. At 0.8B the difference is +2.12 [$-$1.32, +5.84], with unmatched clinical initialization. All intervals are unadjusted patient-bootstrap intervals. The additional full-test comparison of pointer seed 0 against the mean of three fine-tuned LLM seeds gives +0.18 [$-$2.22, +2.39] at 9B; this is not the difference between both systems' three-seed means.

\subsection{Earlier scoring and generation comparison}\label{sub:A.12}

\begin{table}[htbp]
  \centering
  \renewcommand{\thetable}{A11}
  \caption{Earlier evaluation and timing runs, retained for provenance. Accuracy is on 6,975 pairs; latency ratios use 992 pairs and the A/B-logit control as reference, rather than TrialJev as in Table 3. The rows use seed-0 checkpoints. The 9B prompted model is instruct, not the adapted systems' Base checkpoint.}
  \label{tab:A11}
  \footnotesize
  \begin{tabularx}{\linewidth}{P{2.13}P{0.85}P{0.55}P{0.93}P{0.62}P{0.92}}
    \toprule
    \textbf{System} & \textbf{9B accuracy} & \textbf{9B ratio} & \textbf{27B accuracy} & \textbf{27B ratio} & \textbf{27B AUROC} \\
    \midrule
    TrialJev, one-pass & 78.49\% & 0.90 & 81.00\% & 0.82 & 0.885 \\
    TrialJev, two-pass & 78.52\% & 1.54 & 80.97\% & 1.49 & 0.885 \\
    A/B-logit control & 80.50\% & 1.00 & 81.25\% & 1.00 & 0.884 \\
    Zero-shot letters, matched prompt & 43.68\% & --- & 64.00\% & --- & 0.860 \\
    Zero-shot chat, writes the word & 77.35\% & 2.13 & 72.89\% & 1.63 & --- \\
    \bottomrule
  \end{tabularx}
\end{table}

At 27B, the control exceeds prompted word writing by 8.36 points [+3.87, +14.40], or 10.28 with option order swapped. This combines adaptation, input formatting and answer readout. The identical-prompt comparison instead gives +17.25 points [+10.77, +24.41], with AUROC increasing by 0.024 [+0.006, +0.043]. The un-adapted model predicts eligible for 24.4\% of pairs, compared with 56.5\% in the gold labels. This suggests an operating-point contribution but does not measure the accuracy achievable after development-set threshold tuning.

Earlier one-pass timing ratios of 0.82--0.90 and two-pass ratios of 1.49--1.54 are relative to the A/B-logit control in these runs. They are not substituted into the final generation-suite comparison. The instability of absolute latency and the interleaved pointer-path measurements are documented in Appendix A.8.

\subsection{Hosted Jev: calibration, cutoffs and repeat calls}\label{sub:A.13}

\begin{figure}[htbp]
  \centering
  \includegraphics[width=\linewidth]{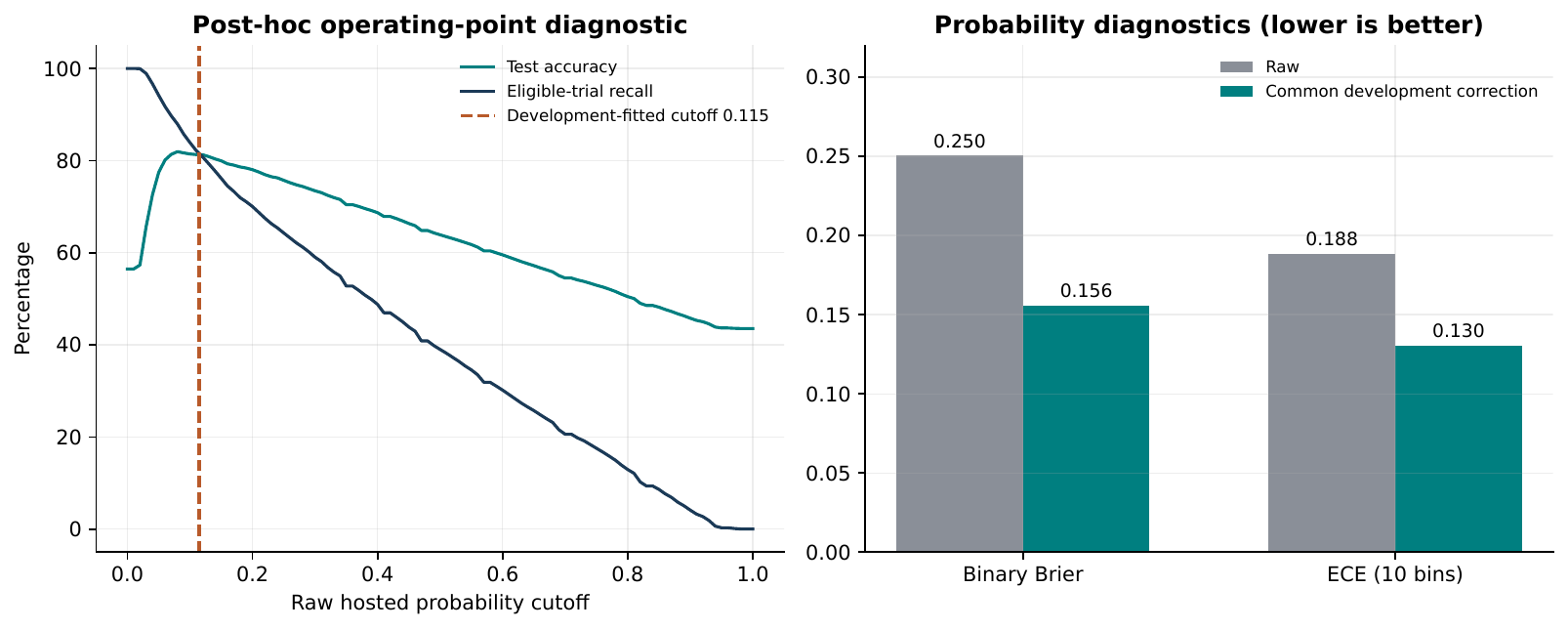}
  \renewcommand{\thefigure}{A3}
  \caption{Hosted Jev operating-point sensitivity on 6,975 test pairs. Left: descriptive accuracy and eligible-trial recall across fixed cutoffs. The highlighted raw-score cutoff, approximately 0.115, follows from the development-fitted map, not selection on this test curve. Right: binary Brier and confidence-bin ECE before and after the common correction. Improved accuracy does not establish calibrated probabilities.}
  \label{fig:A3}
\end{figure}

Hosted Jev's development set contains 2,233 pairs (52.3\% eligible), with raw accuracy 63.01\% and AUROC 0.817; test values are 63.89\% and 0.882. The common development-only fit selects temperature 1.9 and full prior mixing, with class weights 0.655 for excluded and 1.921 for eligible. Development corrected accuracy is 75.55\%; the frozen map yields 81.28\% test accuracy and an effective raw cutoff near 0.115. The prior is derived from development labels, as for all systems in Table 5, not training labels. Binary Brier changes from 0.2505 to 0.1558 and ECE from 0.1885 to 0.1303; these are the probability-quality values for this same common map. A prior historical fit used different parameters and is not the map reported in the corrected comparison. The monotonic map preserves ordering apart from clipping-induced ties. Hosted raw accuracy uses p $\geq$ 0.5: counting an exact tie as excluded instead gives 63.47\%, since API probabilities are rounded to two decimals. Corrected decisions use excluded-on-tie consistently with development fitting.

\textit{Repeat calls.} An overlap between two copies of our job sent 2,638 test pairs to the API twice. For 1,384 of them the two probabilities differed (mean absolute difference 0.018 among changed responses, maximum 0.11), and 23 decisions at the 0.5 cutoff changed. All reported results use one valid answer per pair, the first one in the cleaned file.

\subsection{Exact hosted-Jev paired comparisons}\label{sub:A.14}

\begin{table}[htbp]
  \centering
  \renewcommand{\thetable}{A12}
  \caption{TrialJev versus hosted Jev on the same 6,975 pairs. The displayed TrialJev accuracy column is raw; corrected TrialJev accuracies are in Table 5. Paired differences compare both raw native decisions or both development-corrected decisions, with unadjusted 95\% patient-bootstrap intervals. Three-seed rows average per-pair correctness over the observed seeds. Temperature and prior-mixing strength are selected on development data using its empirical label prior, then frozen. The raw and corrected difference columns therefore use different operating points. Hosted request latency and local inference latency are reported separately in the text.}
  \label{tab:A12}
  \footnotesize
  \begin{tabularx}{\linewidth}{P{1.03}P{1.07}P{0.62}P{0.84}P{1.67}P{0.77}}
    \toprule
    \textbf{TrialJev model} & Raw TrialJev accuracy & \textbf{AUROC} & Both raw: TrialJev minus Jev (63.89\%) & Both corrected: TrialJev minus Jev (81.28\%) & \textbf{AUROC minus Jev (0.882)} \\
    \midrule
    0.8B, fresh (3 seeds) & 54.27 $\pm$ 0.95\% & 0.582 & $-$9.62 [$-$14.16, $-$4.66] & $-$27.38 [$-$31.28, $-$23.39] & $-$0.300 \\
    0.8B, MedJev init (3 seeds) & 70.71 $\pm$ 2.64\% & 0.786 & +6.82 [+1.33, +12.74] & $-$10.47 [$-$13.22, $-$7.71] & $-$0.095 \\
    9B (3 seeds) & 78.27 $\pm$ 0.55\% & 0.862 & +14.39 [+8.98, +20.19] & $-$2.34 [$-$3.60, $-$0.95] & $-$0.020 \\
    27B & 80.97\% & 0.885 & +17.09 [+10.88, +23.66] & $-$0.29 [$-$1.16, +0.62] & +0.003 \\
    Qwen3.8-27B & 81.84\% & 0.888 & +17.95 [+12.08, +24.60] & +0.60 [$-$0.30, +1.49] & +0.006 \\
    \bottomrule
  \end{tabularx}
\end{table}

\begin{figure}[htbp]
  \centering
  \includegraphics[width=\linewidth]{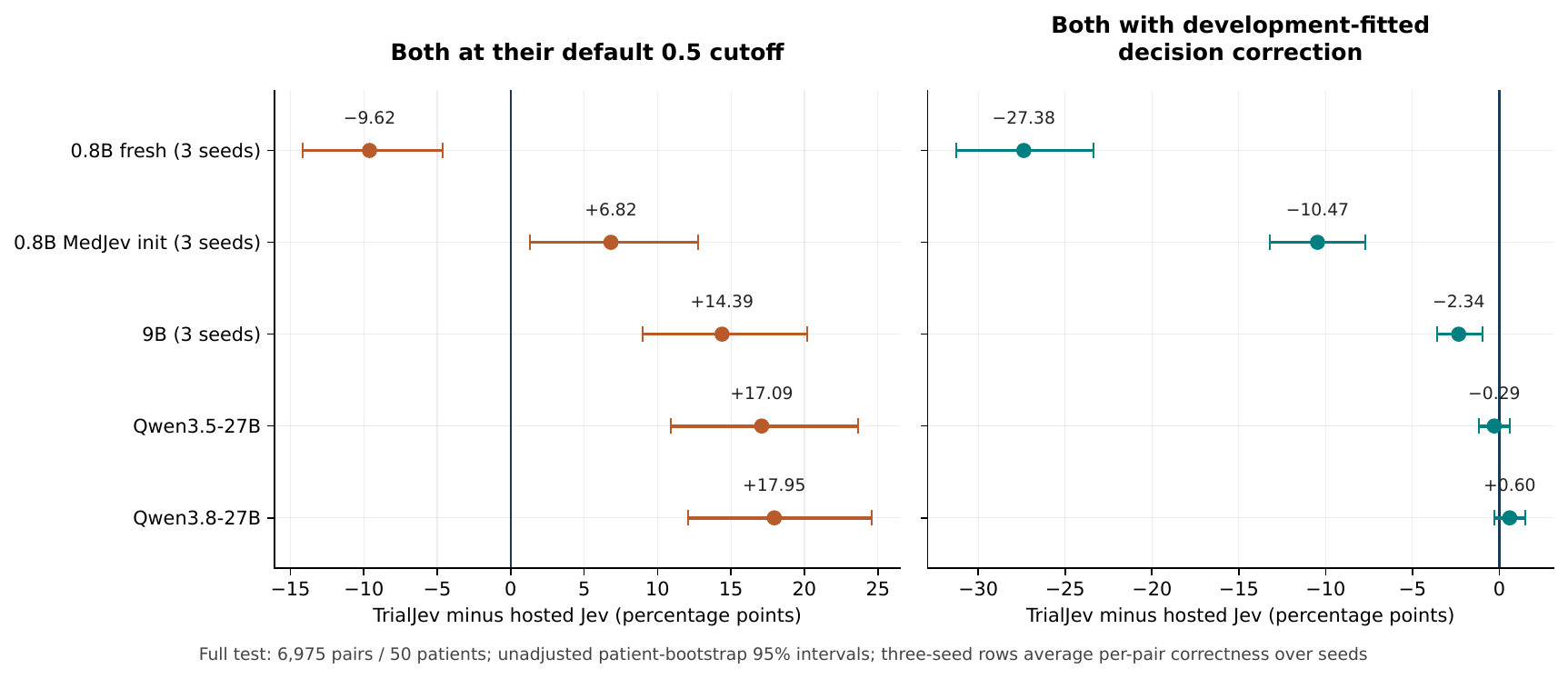}
  \renewcommand{\thefigure}{A4}
  \caption{Paired full-test accuracy differences, TrialJev minus hosted Jev, on 6,975 pairs from 50 patients. Points give differences and bars give unadjusted 95\% patient-bootstrap intervals. Left: both systems at their default 0.5 cutoff. Right: the same development-fitted correction applied to both TrialJev and Jev. Three-seed TrialJev rows average per-pair correctness across the observed seeds. An interval crossing zero does not establish equivalence. Exact differences are in Table A12.}
  \label{fig:A4}
\end{figure}

Against raw Jev, TrialJev is more accurate at every size except the fresh 0.8B model, by 6.8 to 18.0 points. With the same development-fitted correction applied to both, the paired accuracy intervals include zero at 27B and on Qwen3.8-27B; this is similar observed accuracy, not evidence of equivalence. TrialJev is below corrected Jev at 9B (by 2.3 points) and at 0.8B (by 10.5 points). Jev's AUROC (0.882) is numerically close to the 27B models and exceeds the 9B model by 0.020. We do not know why Jev ranks this well: it is hosted, its size and training data are not disclosed, and we cannot exclude exposure to related text. Jev was therefore used without adaptation by us, not shown to be untrained on TREC. Repeat calls did not always return identical probabilities: of the pairs we accidentally requested twice, half returned a slightly different probability (mean absolute difference 0.018 among the changed responses) and 0.9\% crossed the 0.5 cutoff, and probabilities are returned rounded to two decimals (Appendix A.13). Our results use one call per pair. The comparison does not show that TrialJev is more accurate than calibrated Jev; its supported practical distinction is local, open-weight inference. The hosted latency is a measured round-trip time on unknown service hardware, not a controlled hardware comparison.

\subsection{Full accuracy and serving-path comparison}\label{sub:A.15}

\begin{table}[H]
  \centering
  \renewcommand{\thetable}{A13}
  \caption{TrialJev against a fine-tuned LLM. Full-test columns use 6,975 pairs and the canonical two-pass evaluator for TrialJev; the one-pass and two-pass rows have distinct measured timing-sample accuracies where predictions differ (full-test accuracy of the one-pass path changes by at most 0.09 points, Table A6; the 9B timing subset differs by 0.50 points). Timing columns use 992 identical pairs per size, seed-0 checkpoints, batch size 1, one GPU, systems run one after another (H100 NVL/bf16 at 9B to 27B, RTX 6000 Ada/fp32 at 0.8B); the ratio is median time divided by that of TrialJev's one-pass path in the same chain. Mean $\pm$ SD is given for three-seed configurations, with seed 0 in parentheses. The 0.8B TrialJev starts from the clinical initialization (fresh 0.8B TrialJev: 54.27 $\pm$ 0.95\%), whereas the 0.8B fine-tuned LLM starts fresh.}
  \label{tab:A13}
  \footnotesize
  \setlength{\tabcolsep}{4pt}
  \begin{tabularx}{\linewidth}{P{1.04}P{1.44}P{1.30}P{0.99}P{1.08}P{0.72}P{0.43}}
    \toprule
    \textbf{Size} & \textbf{System} & \textbf{Full-test accuracy} & \textbf{Full-test AUROC} & \textbf{Accuracy, n = 992} & \textbf{Median ms} & \textbf{Ratio} \\
    \midrule
    0.8B & \textbf{TrialJev, one-pass} & 70.71 $\pm$ 2.64 (seed 0: 73.48) & 0.786 & 74.80\% & 54.0 & 1.00 \\
    0.8B & TrialJev, two-pass & same & same & 74.80\% & 69.2 & 1.28 \\
    0.8B & Fine-tuned LLM & 74.05\% & 0.802 & 72.68\% & 99.9 & 1.85 \\
    9B & \textbf{TrialJev, one-pass} & 78.27 $\pm$ 0.55 (seed 0: 78.52) & 0.862 & 80.85\% & 106.2 & 1.00 \\
    9B & TrialJev, two-pass & same & same & 80.34\% & 183.2 & 1.73 \\
    9B & Fine-tuned LLM & 78.34 $\pm$ 2.31 (seed 0: 79.50) & 0.862 & 80.65\% & 259.9 & 2.45 \\
    27B & \textbf{TrialJev, one-pass} & 80.97\% & 0.885 & 81.25\% & 177.6 & 1.00 \\
    27B & TrialJev, two-pass & same & same & 81.25\% & 328.3 & 1.85 \\
    27B & Fine-tuned LLM & 79.87\% & 0.879 & 80.14\% & 464.4 & 2.62 \\
    Qwen3.8-27B & \textbf{TrialJev, one-pass} & 81.84\% & 0.888 & 82.66\% & 178.2 & 1.00 \\
    Qwen3.8-27B & TrialJev, two-pass & same & same & 82.66\% & 330.1 & 1.85 \\
    Qwen3.8-27B & Fine-tuned LLM & 79.93\% & 0.888 & 80.54\% & 480.3 & 2.70 \\
    \bottomrule
  \end{tabularx}
\end{table}

\subsection{Zero-shot explain-then-answer baseline}\label{sub:A.16}

\begin{table}[htbp]
  \centering
  \renewcommand{\thetable}{A14a}
  \caption{Explain-then-answer accuracy on 200 identical pairs per system, four per patient. This is zero-shot explanation with native thinking disabled and a 512-token cap. Invalid output counts as wrong; every invalid output reached the cap. The valid-only column uses fewer pairs and is not directly comparable with the all-pair columns. The last column is TrialJev minus explanation, in percentage points with an unadjusted patient-bootstrap interval. No 0.8B explanation condition was evaluated.}
  \label{tab:A14a}
  \footnotesize
  \setlength{\tabcolsep}{4pt}
  \begin{tabularx}{\linewidth}{P{1.04}P{0.95}P{1.21}P{0.65}P{0.65}P{0.95}P{1.55}}
    \toprule
    \textbf{Size} & \textbf{Explanation} & \textbf{Valid only (n)} & \textbf{Invalid / 200} & \textbf{TrialJev} & \textbf{Fine-tuned LLM} & \textbf{Gap [95\% CI]} \\
    \midrule
    9B & 65.5\% & 72.8\% (n = 180) & 20 & 83.0\% & 81.0\% & +17.5 [+11.5, +24.0] \\
    27B & 61.5\% & 71.1\% (n = 173) & 27 & 79.5\% & 83.0\% & +18.0 [+10.0, +26.0] \\
    Qwen3.8-27B & 52.5\% & 75.5\% (n = 139) & 61 & 82.0\% & 81.0\% & +29.5 [+22.0, +37.5] \\
    \bottomrule
  \end{tabularx}
\end{table}

\begin{table}[htbp]
  \centering
  \renewcommand{\thetable}{A14b}
  \caption{Zero-shot explanation baseline, time, on the same 200 pairs (batch 1). Ratio = median time of the explanation divided by the median time of TrialJev's one-pass path on the same pairs.}
  \label{tab:A14b}
  \footnotesize
  \begin{tabularx}{\linewidth}{P{0.58}P{0.81}P{1.40}P{1.21}}
    \toprule
    \textbf{Size} & \textbf{Mean new tokens} & \textbf{Explanation, median s per pair} & \textbf{Ratio to TrialJev one-pass} \\
    \midrule
    9B & 270 & 7.3 & 68 \\
    27B & 312 & 19.2 & 104 \\
    Qwen3.8-27B & 383 & 25.4 & 139 \\
    \bottomrule
  \end{tabularx}
\end{table}

\textit{The cost of explanation.} Under this prompt and 512-token budget, explain-then-answer generation has median latency 68--139 times TrialJev's and strict accuracy 17.5--29.5 points lower on the same 200 pairs. All three paired intervals exclude zero. Twenty, 27 and 61 outputs lack a valid verdict and reach the token cap. The baseline's valid-only accuracy is 71.1--75.5\%; on those same valid-output subsets, TrialJev scores 83.9\%, 80.3\% and 84.2\%, respectively. This separates verdict failure from accuracy among completed answers, but conditioning on valid outputs is itself selective. The experiment measures a capped zero-shot explanation strategy, not an inherent disadvantage of reasoning. Longer-budget or fine-tuned reasoning and explanation quality were not evaluated.

\subsection{Adaptation and serving-path effects}\label{sub:A.17}

On the identical lettered prompt, the un-adapted Qwen3.5-27B backbone scores 64.00\%, versus 81.25\% after A/B-logit adaptation: +17.25 points [+10.77, +24.41]. Its AUROC rises from 0.860 to 0.884. The much larger change in thresholded accuracy than ranking is consistent with an operating-point or letter-bias effect; no threshold was tuned to test that explanation. Comparing prompted word writing instead gives 72.89\%, but changes both prompt and readout and therefore cannot isolate adaptation (Appendix A.7).

Serving-path selection is equally consequential for latency. The one-pass and cached two-pass implementations do not give identical floating-point results. On the full test set at 9B--27B they disagree on 12--20 pairs (0.17--0.29\%), with accuracy changes of at most 0.09 percentage points. The 9B timing subset changes by 0.50 points. We retain canonical full-test results and identify the one-pass timing reference explicitly rather than treating the paths as interchangeable. Earlier timing runs and their variation are retained in Appendices A.8 and A.12.

\section{Methods in detail}\label{app:B}

\subsection{Models and starting points}\label{sub:B.1}

\begin{table}[htbp]
  \centering
  \renewcommand{\thetable}{B1}
  \caption{Backbones. All are Qwen hybrid models (Gated DeltaNet layers, via flash-linear-attention). ``Revision'' is the Hub commit used.}
  \label{tab:B1}
  \footnotesize
  \begin{tabularx}{\linewidth}{P{1.09}P{0.42}P{0.91}P{1.00}P{1.58}}
    \toprule
    \textbf{Backbone} & \textbf{Params} & \textbf{Type} & \textbf{Revision} & \textbf{Readouts trained} \\
    \midrule
    Qwen3.5-0.8B-Base & 0.8B & pretraining-only & not recorded & pointer (3 seeds fresh, 3 seeds MedJev-init), A/B-logit LoRA (1 seed), fine-tuned LLM (1 seed), zero-shot plain and zero-shot letters \\
    Qwen3.5-9B-Base & 9B & pretraining-only & \texttt{68c46c4b\ldots{}f404}\ddag{} & pointer (3 seeds), A/B-logit LoRA (3 seeds), fine-tuned LLM (3 seeds), zero-shot plain and zero-shot letters \\
    Qwen3.5-9B & 9B & post-trained (instruct) & \texttt{c2022362\ldots{}7b9a} & zero-shot chat and zero-shot explanation baselines only \\
    Qwen3.5-27B & 27B & post-trained backbone & \texttt{fc05daec\ldots{}f654}\ddag{} & pointer, A/B-logit LoRA, fine-tuned LLM, zero-shot chat, zero-shot explanation and zero-shot letters \\
    Qwen3.8-27B & 27B & post-trained backbone & \texttt{1d4bf0f2\ldots{}0c0} & pointer, A/B-logit LoRA, fine-tuned LLM, zero-shot chat and zero-shot explanation \\
    \bottomrule
  \end{tabularx}

  \smallskip
  \raggedright\footnotesize \ddag{} Taken from the only snapshot in the training cache; the seed-0 pointer-head checkpoints record no revision. For the LM-logit runs and the later pointer-head seeds at 9B the revision was passed explicitly. The generation runs and the full-set pointer-head timing runs used the revisions listed (the cached snapshot is recorded with each pointer-head timing run).
\end{table}

``Fresh'' means newly initialized LoRA weights, plus a newly initialized pointer head for that readout. ``Clinical initialization'' starts the 0.8B pointer-head model from the released MedJev adapter and head. We did not initialize the control or fine-tuned LLM from the clinical adapter alone; the absence of a compatible released pointer head for those systems does not make adapter-only transfer impossible.

\subsection{Input construction}\label{sub:B.2}

Both readouts see the same text. The patient note (the state) is tokenized and truncated from the head at 511 tokens. The question reads ``Is this patient eligible for the clinical trial below? Eligible means every inclusion criterion is met and no exclusion criterion applies.'', followed by the trial text. Corpus text is rewritten so that it cannot produce a reserved delimiter token. The two options are ``Excluded: an inclusion criterion is not met or an exclusion criterion applies'' and ``Eligible: meets the inclusion criteria and no exclusion criterion applies''.

\subsection{Readout 1: the pointer head}\label{sub:B.3}

Each question is rendered as the state followed by \texttt{[question] [option 1] </opt> [option 2] </opt> <decide>}, using rarely used Qwen special tokens as delimiters. A bilinear pointer head of dimension 256 compares the hidden state at \texttt{<decide>} with the hidden state at the end of each option span, and a softmax over the two scores gives the probabilities. Only the LoRA weights and the head are trained. On hybrid backbones the state is recomputed for each question during training. At serving time the state can be encoded once and each question reads its cache.

\subsection{Readout 2: the A/B-logit control}\label{sub:B.4}

The prompt is the state, the question and trial, then \texttt{A. <first option>}, \texttt{B. <second option>} and \texttt{Answer:} (Appendix D.1). The probability is a softmax over the language-model head's logits for the tokens `` A'' and `` B'' at the last position. No parameters are added beyond LoRA. Training minimizes cross-entropy over the two letter logits, matching the pointer head's two-way cross-entropy. This is not conventional generative fine-tuning, which trains the model to write an answer sequence with a cross-entropy over the whole vocabulary at each answer position. Here the two label-token scores at one position are renormalized against each other and a binary decision is optimized; nothing else is supervised, so the adapter is trained only to separate the two letters. The setup follows prompt-based fine-tuning with a verbalizer \cite{ref7} and the lettered multiple-choice prompt format \cite{ref8}, applied to a causal language model. The default order is A = Excluded, B = Eligible. A swapped variant (A = Eligible) is scored as a check on letter bias.

\subsection{Training}\label{sub:B.5}

Both readouts use LoRA of rank 16 with $\alpha$ = 32 and dropout 0.05 on the attention, MLP and (for hybrid models) Gated DeltaNet projection groups. Optimization is AdamW with weight decay 0.01, learning rate 1e-4 on a one-cycle schedule with 5\% warm-up, gradient clipping at 1.0, three epochs (1,746 optimizer steps), effective batch 16, gradient checkpointing and bf16 autocast. The seed is 0 unless stated; seeds 1 and 2 at 9B (both readouts) were run with the same command and a different \texttt{-{}-seed}. The LM-logit script reproduces the pointer-head trainer's first-epoch record order for a given seed, its epoch reshuffling rule, its validation on the full development split and its log format. The 0.8B runs use an fp32 backbone, batch 4 $\times$ accumulation 4 and validation every 100 steps. The 9B and 27B runs use a bf16 backbone, batch 2 $\times$ accumulation 8 and validation every 200 steps.

\subsection{Checkpoint selection and test protocol}\label{sub:B.6}

Checkpoints are saved periodically, and the one with the lowest development loss is evaluated once on the test set. The development loss is a proper scoring rule, so it favors well-scored probabilities. Departures from this protocol occurred and are reported where they matter: the seed-0 9B LM-logit model, the 27B LM-logit model and the Qwen3.8-27B LM-logit model were each scored a second time with the option letters swapped, as a diagnostic and not for selection. The rule is applied unchanged to every run; it selected step 200 for two LM-logit runs (9B seed 2 and Qwen3.8-27B) and step 1000 for one pointer-head run (9B seed 2) (Appendix A.6). After the main analysis, the step-400 checkpoints of the two LM-logit runs that the rule had sent to step 200 (9B seed 2 and Qwen3.8-27B) were also scored on the test set, as a labeled post-hoc check; they are reported separately in Appendix A.6, Table A3, and not used for any selection.

\subsection{Generation baselines}\label{sub:B.7}

Earlier full-test experiments measured prompted word writing, emission of the A/B classifier's label token, letter-logit scoring and matched-prompt zero-shot scoring. Writing uses greedy decoding and strict parsing; scoring does not decode. The 9B--27B runs use batch size 1 and bf16, with up to eight new tokens for prompted word writing and one for A/B-label emission. The separate 0.8B runs use fp32 inference.

\begin{itemize}
  \item \textit{Zero-shot chat.} The post-trained model (Qwen3.5-9B, Qwen3.5-27B, Qwen3.8-27B) gets its chat template with thinking disabled, a one-sentence system instruction, the patient note, the question and trial text, the two options and the line ``Reply with exactly one word: Eligible or Excluded.'' (Appendix D.2). It has no adapter.
  \item \textit{Zero-shot plain.} Qwen3.5-9B-Base gets the same user text followed by ``Answer:'', with no chat template.
  \item \textit{A/B-logit LoRA, emits its label token.} The adapter gets its exact training prompt (Appendix D.1) and generates one token, which must be `` A'' or `` B''. This is the same classifier allowed to output its selected label token; it is not conventional generative fine-tuning.
  \item \textit{A/B-logit LoRA, reads the logits.} The same adapter and prompt, with the answer read from the two letter logits in one pass (Appendix B.4).
  \item \textit{Zero-shot letters (matched prompt).} The un-fine-tuned backbone (Qwen3.5-0.8B-Base, Qwen3.5-9B-Base, Qwen3.5-27B), with no adapter, gets the exact fine-tuning prompt of Appendix D.1, and the answer is read from the `` A'' and `` B'' logits in one pass. It differs from the A/B-logit classifier only by the adapter, so it measures what fine-tuning adds on the same prompt. It was run after the main analysis.
\end{itemize}

Both option orders were run for the chat, fine-tuned and matched-prompt conditions; the Base word prompt was run in one order. Parsing is strict: the output, after stripping punctuation and case, must be exactly one of the two permitted words (or letters). Anything else is invalid, counts as wrong in accuracy, and counts as the opposite of the gold class in macro-F1. There is no fallback to the logits. Two limits apply to reading these results. The 9B zero-shot model is the instruct checkpoint, whereas the fine-tuned 9B model starts from Base, so the 9B gap is not a pure effect of fine-tuning; at 27B and for Qwen3.8 the zero-shot model and the backbone are the same checkpoint. And the fine-tuned adapters were trained with a loss over two letters (Appendix B.4), so the ``emits its label token'' condition is a one-token readout and says nothing about a conventionally fine-tuned generative model; that is why the fine-tuned LLM described below was added. The comparison of the word-writing zero-shot conditions with the fine-tuned ones combines the effect of fine-tuning with that of the prompt and answer format; the matched-prompt condition separates them.

Two further conditions were added after the main analysis (Section 4.3). \textit{Fine-tuned LLM}: the adapter trained on the zero-shot word prompt, with the target `` Eligible'' or `` Excluded'' plus the end-of-sequence token and a full-vocabulary cross-entropy on the answer tokens only. It is evaluated on all 6,975 pairs with greedy decoding until the end token (at most 8 new tokens), strict parsing and the excluded-first option order only; the two-way score is the softmax over the first-step scores of the first tokens of the two words. \textit{Zero-shot explanation} (\texttt{zeroshot-reason}): the instruct model with its chat template and thinking disabled, greedy decoding up to 512 new tokens, and the verdict taken from the last line that matches ``Answer: Eligible'' or ``Answer: Excluded'' (case-insensitive); output without one is invalid. It was run on the 200-pair subset only. The parser in the released script takes the last match anywhere in the text; a stricter check that requires the match on the final non-empty line gave the same verdict for all 600 saved outputs (200 pairs at each of three sizes).

\subsection{Metrics and statistics}\label{sub:B.8}

We report accuracy, macro-F1, AUROC, conventional binary Brier score and ECE in ten equal-width confidence bins. Binary Brier is half the evaluator's two-class sum. Bootstrap intervals resample the 50 patients with replacement 10,000 times and use the 2.5th and 97.5th percentiles. All pairs belonging to each selected patient are retained, including repeated patient selections, so the estimand is pair-weighted accuracy with uncertainty clustered by patient. Differences use identical resamples. Hosted-Jev AUROC difference intervals use 2,000 resamples; its paired accuracy intervals use 10,000. These intervals condition on the fitted decision-correction parameters, observed training seeds and saved API responses; they do not include uncertainty from development fitting or repeat API calls. The reported two-sided bootstrap tail measure is min(1, 2 $\times$ min\{Pr($\Delta$ $\leq$ 0), Pr($\Delta$ $\geq$ 0)\}); comparisons are unadjusted and exploratory, not formal equivalence tests. Multi-seed metrics average the separately evaluated seeds, with sample SD reported separately; patient resampling does not include seed uncertainty. The common correction searches temperatures 0.5--3.0 in increments of 0.1 and prior-mixing strengths 0, 0.25, 0.5, 0.75 and 1.0, using development accuracy; tied grid optima retain the first maximum. The class weights use the development label prior and mean development prediction. Both fitting and corrected test evaluation apply argmax to [P(excluded), P(eligible)], resolving an exact tie as excluded. Raw saved decisions retain their native conventions, including p $\geq$ 0.5 for hosted Jev and the fine-tuned LLM's auxiliary score. This convention matters for exact ties and is kept explicit rather than silently changing the reported native results.

Generated verdicts use strict accuracy, invalid-answer counts and the evaluator's strict macro-F1 convention, which treats an invalid answer as the opposite of its gold class. This is a custom failure-penalized binary F1, not a three-class abstention metric. The fine-tuned LLM's first-token two-way score is used for auxiliary AUROC and the raw and development-corrected score-based decision comparisons in Table 5. It is neither the probability of the complete generated sequence nor a validated calibrated eligibility probability. Generated-verdict accuracy uses the decoded output, not that score. Option-swapped and ensemble analyses are diagnostics, not checkpoint-selection criteria.

\subsection{Efficiency measurement}\label{sub:B.9}

The per-pair timing window includes the scoring or generation call and GPU synchronization, but excludes model loading, tokenization, input preparation and text parsing. Warm-up precedes measurement. We report the ratio of median latencies, not the median of pairwise ratios unless explicitly labeled. A median cannot be multiplied by the number of candidates to establish measured total screening time; we make no whole-pool or staff-time extrapolation.

The final suite uses 992 pairs, selected by taking up to 20 per patient in SHA-256 identifier order. Each chain runs control, fine-tuned LLM, both TrialJev paths and prompted one-word generation sequentially on one GPU, followed by explanation on four pairs per patient. Every reasoning comparison uses those identical 200 pairs for both accuracy and its TrialJev latency denominator. The 9B, 27B and Qwen3.8 chains run concurrently on separate H100 NVL cards. The 0.8B chain runs alone on RTX 6000 Ada with fp32 inference. Chain wall times are 0.1, 0.8, 1.6 and 1.8 hours for 0.8B, 9B, 27B and Qwen3.8-27B.

Pointer paths are warmed and timed in the same process with alternating A/B order per pair. Different systems are sequential, not interleaved or randomly ordered. The fine-tuned LLM's timed Hugging Face call also collects generation scores; its three emitted tokens include EOS. These implementation choices, shared-node load and one chain per size limit causal attribution of the latency ratio to architecture alone. The subsequent full-test timing distributed conditions across four cards. A quiet-node check on identical 200 pairs gave 111.6, 113.9, 114.0 and 113.0 ms across cards, but does not establish stable timing for every condition. Earlier timing variation is retained in Appendix A.8.

\subsection{Hosted Jev API}\label{sub:B.10}

Requests go to \texttt{https://api.typesafe.ai/v1/systemone} with model name \texttt{jev-latest}; every response reported the model string \texttt{jev-1.13.0}. The request is the processed record with its label removed: the patient note as the state and one typed yes/no question holding the trial text and the two options. The answer is the probability for ``true''. The client accepts a response only if the answer for the eligible question is a finite number in [0, 1], saves the returned model string and the time with every row, stops if the model string changes or differs from the pinned \texttt{jev-1.13.0}, refuses to resume a file whose ids or gold labels do not match the data, retries server and rate-limit errors and recoverable request failures, and records both the latency of the successful attempt and the elapsed time including retries. Its API key is read from the environment or a local key file outside the repository and is never printed or saved. Only synthetic TREC patient notes and public registry text were sent.

Runs: a 20-pair pilot; the 992-pair timing sample, one request at a time (992 valid answers, no retries; a first attempt overlapped with a second copy of the job and was set aside, then re-run alone); the development split (2,233 pairs) and the full test split (6,975 pairs) with six parallel requests (about 18 pairs per second in the clean development run). The test file was assembled from the overlapped run by dropping 12 corrupted lines and duplicate rows and keeping one valid answer per pair; the raw file is kept. The processed requests contained about 1,140 input tokens per pair.  

\section{Training configuration and cost}\label{app:C}

\begin{table}[htbp]
  \centering
  \renewcommand{\thetable}{C1}
  \caption{Settings per group of runs (both readouts unless noted).}
  \label{tab:C1}
  \footnotesize
  \begin{tabularx}{\linewidth}{P{1.00}P{1.03}P{0.97}}
    \toprule
    \textbf{Setting} & \textbf{0.8B runs} & \textbf{9B / 27B / Qwen3.8-27B runs} \\
    \midrule
    LoRA rank / $\alpha$ / dropout & 16 / 32 / 0.05 & 16 / 32 / 0.05 \\
    Pointer-head dimension (pointer only) & 256 & 256 \\
    Learning rate / schedule & 1e-4, one-cycle (5\% warm-up) & same \\
    Weight decay / optimizer & 0.01, AdamW & same \\
    Gradient clipping & 1.0 & 1.0 \\
    Epochs / optimizer steps & 3 / 1,746 & 3 / 1,746 \\
    Effective batch & 16 (4 $\times$ 4) & 16 (2 $\times$ 8) \\
    Token budgets (state / total) & 512 / 4,096 & 512 / 4,096 \\
    Backbone precision / autocast & fp32 / bf16 & bf16 / bf16 \\
    Validation cadence & every 100 steps & every 200 steps \\
    Seeds & 0 (all configurations); 1, 2 (pointer head) & 0; 1 and 2 also at 9B for both readouts \\
    Selection & lowest development loss & lowest development loss \\
    \bottomrule
  \end{tabularx}
\end{table}

\begin{table}[htbp]
  \centering
  \renewcommand{\thetable}{C2}
  \caption{Development loss and accuracy at the selected step (15 development patients, 2,233 pairs), with training wall time and peak GPU memory (one H100 NVL for 9B/27B; one RTX 6000 Ada for 0.8B; shared machines).}
  \label{tab:C2}
  \footnotesize
  \begin{tabularx}{\linewidth}{P{1.98}P{0.83}P{0.57}P{0.87}P{0.89}P{0.86}}
    \toprule
    \textbf{Run} & \textbf{Selected step} & \textbf{Dev loss} & \textbf{Dev accuracy} & \textbf{Train wall time} & \textbf{Peak memory} \\
    \midrule
    0.8B pointer, fresh, seed 0 & 1000 & 0.6747 & 59.11\% & 2.36 h & 7.2 GB \\
    0.8B pointer, fresh, seed 1 & 100 & 0.6883 & 53.83\% & 2.30 h & 7.2 GB \\
    0.8B pointer, fresh, seed 2 & 500 & 0.7224 & 54.86\% & 2.32 h & 7.2 GB \\
    0.8B pointer, MedJev init, seed 0 & 400 & 0.5727 & 70.22\% & 2.31 h & 7.2 GB \\
    0.8B pointer, MedJev init, seed 1 & 500 & 0.5828 & 69.55\% & 2.32 h & 7.2 GB \\
    0.8B pointer, MedJev init, seed 2 & 500 & 0.5831 & 70.31\% & 2.30 h & 7.2 GB \\
    0.8B A/B-logit LoRA, seed 0 & 200 & 0.6903 & 54.14\% & 1.73 h & 7.2 GB \\
    9B pointer, seed 0 & 400 & 0.5077 & 77.70\% & 3.77 h & 21.2 GB \\
    9B pointer, seed 1 & 400 & 0.5163 & 74.97\% & 3.33 h & 21.2 GB \\
    9B pointer, seed 2 & 1000 & 0.5417 & 76.31\% & 3.37 h & 21.2 GB \\
    9B A/B-logit LoRA, seed 0 & 400 & 0.4941 & 76.71\% & 3.01 h & 23.0 GB \\
    9B A/B-logit LoRA, seed 1 & 400 & 0.5080 & 77.65\% & 2.77 h & 23.0 GB \\
    9B A/B-logit LoRA, seed 2 & 200 & 0.5176 & 76.53\% & 2.80 h & 23.0 GB \\
    27B pointer & 400 & 0.4871 & 79.00\% & 9.31 h & 59.6 GB \\
    27B A/B-logit LoRA & 400 & 0.4940 & 78.10\% & 8.08 h & 61.9 GB \\
    Qwen3.8-27B pointer & 400 & 0.4918 & 77.88\% & 9.52 h & 59.6 GB \\
    Qwen3.8-27B A/B-logit LoRA & 200 & 0.5130 & 75.68\% & 8.06 h & 61.9 GB \\
    \bottomrule
  \end{tabularx}
\end{table}

\begin{table}[htbp]
  \centering
  \renewcommand{\thetable}{C3}
  \caption{Fine-tuned LLM: selected step, development loss and first-token accuracy at that step (15 development patients, 2,233 pairs), training wall time and logged peak GPU memory. The development loss is the cross-entropy on the answer tokens (the word and the end token) and is not comparable with the losses in Table C2; the accuracy is that of the first answer token. Settings are those of Table C1, with a validation and checkpoint cadence of 100 steps (0.8B) or 200 steps (others) and the selection rule unchanged.}
  \label{tab:C3}
  \footnotesize
  \begin{tabularx}{\linewidth}{P{1.88}P{0.77}P{0.54}P{1.26}P{0.76}P{0.79}}
    \toprule
    \textbf{Run} & \textbf{Selected step} & \textbf{Dev loss} & \textbf{Dev first-token accuracy} & \textbf{Train wall time} & \textbf{Peak memory} \\
    \midrule
    0.8B fine-tuned LLM, fresh, seed 0 & 800 & 0.2196 & 69.91\% & 1.79 h & 7.2 GB \\
    9B fine-tuned LLM, seed 0 & 400 & 0.1750 & 73.94\% & 2.79 h & 23.1 GB \\
    9B fine-tuned LLM, seed 1 & 400 & 0.1700 & 74.79\% & 2.81 h & 23.1 GB \\
    9B fine-tuned LLM, seed 2 & 200 & 0.1675 & 76.67\% & 2.85 h & 23.1 GB \\
    27B fine-tuned LLM & 200 & 0.1779 & 76.58\% & 8.15 h & 61.9 GB \\
    Qwen3.8-27B fine-tuned LLM & 200 & 0.1720 & 76.80\% & 8.18 h & 61.9 GB \\
    \bottomrule
  \end{tabularx}
\end{table}

\section{Prompts}\label{app:D}

\subsection{Prompt for the A/B-logit control}\label{sub:D.1}

\begin{Verbatim}[breaklines,fontsize=\small]
<patient note, truncated to 511 tokens>

Is this patient eligible for the clinical trial below? Eligible means every inclusion criterion is met and no exclusion criterion applies.

<trial text: TITLE, CONDITIONS, INTERVENTIONS, SUMMARY, DEMOGRAPHICS, ELIGIBILITY, ...>

A. Excluded: an inclusion criterion is not met or an exclusion criterion applies
B. Eligible: meets the inclusion criteria and no exclusion criterion applies
Answer:
\end{Verbatim}

The probability of ``eligible'' is the softmax over the logits of `` A'' and `` B'' at the final token, mapped back to the options. In the swapped variant the two option lines exchange letters.

\subsection{Prompts for the generation baselines}\label{sub:D.2}

The user text is shared by both zero-shot conditions; the line ``Options:'' is followed by the two option lines, in the order excluded then eligible (the swapped variant exchanges them).

\begin{Verbatim}[breaklines,fontsize=\small]
Patient note:
<patient note>

Is this patient eligible for the clinical trial below? Eligible means every inclusion criterion is met and no exclusion criterion applies.

<trial text>

Options:
Excluded: an inclusion criterion is not met or an exclusion criterion applies
Eligible: meets the inclusion criteria and no exclusion criterion applies

Reply with exactly one word: Eligible or Excluded.
\end{Verbatim}

\textit{Zero-shot chat:} the Qwen chat template with thinking disabled, the system message ``You are a clinical trial screening assistant. Decide whether the patient is eligible for the clinical trial. Reply with exactly one word: Eligible or Excluded.'' and the text above as the user message; the model generates greedily up to 8 tokens. \textit{Zero-shot plain:} the text above followed by a newline and ``Answer:'', with no template. \textit{A/B-logit control:} the prompt of Appendix D.1, with one generated label token (writing) or one pass over the letter logits (scoring). \textit{Fine-tuned LLM:} the word-answer prompt of Appendix D.3, with greedy decoding until the end-of-sequence token or 8 new tokens. \textit{Zero-shot letters (matched prompt):} the prompt of Appendix D.1 with no adapter, scored from the letter logits.

\subsection{Prompts for the fine-tuned LLM and the explanation baseline}\label{sub:D.3}

\textit{Fine-tuned LLM:} the text of D.2 followed by a newline and ``Answer:'', with no chat template (the same as the zero-shot plain prompt); the training target is `` Eligible'' or `` Excluded'' followed by the end-of-sequence token.

\textit{Zero-shot explanation:} the Qwen chat template with thinking disabled, the system message ``You are a clinical trial screening assistant. Decide whether the patient is eligible for the clinical trial. Reason briefly about the inclusion and exclusion criteria, then give the verdict on the last line as `Answer: Eligible' or `Answer: Excluded'.'' and, as the user message, the patient note, the question and trial text and the two options of D.2 followed by ``Explain your reasoning briefly, then end with exactly one line: Answer: Eligible or Answer: Excluded.'' The model generates greedily up to 512 tokens.

\section{Environments}\label{app:E}

Larger models (9B, 27B, Qwen3.8-27B): 4 $\times$ NVIDIA H100 NVL (95,830 MiB), driver 610.43.02, Python 3.12.9, torch 2.14.0+cu130, transformers 5.17.0, peft 0.21.0, accelerate 1.15.0, flash-linear-attention 0.5.2, numpy 2.5.3. 0.8B models: 4 $\times$ NVIDIA RTX 6000 Ada (49,140 MiB), driver 590.48.01, Python 3.12.14, torch 2.14.0, same transformers, peft, accelerate and flash-linear-attention versions. Base-model revisions are in Table B1; the verdict-fine-tuned LLM runs passed the 9B, 27B and Qwen3.8-27B revisions explicitly. The 9B and 27B generation and timing runs used the larger-model environment; the 0.8B timing and matched-prompt runs used the 0.8B environment (torch 2.14.0+cu130, transformers 5.17.0, peft 0.21.0; the reference implementation of the causal convolution was used because the optimized kernel was not installed). The 992-pair timing sample takes up to 20 pairs per patient in the order of the SHA-256 hash of the pair identifier (all 50 patients); the 9B and 27B chain ran on separate cards of one node. The full-test-set generation run placed the 13 conditions and the three pointer-head timing runs on the four GPUs of the same node at once, each GPU checked as idle before its first condition. The final timing chains (Table 3 and Table A13) used the larger-model environment for 9B, 27B and Qwen3.8-27B and the 0.8B environment (Appendix B.9).

\section{Related-work context}\label{app:F}

\begin{table}[htbp]
  \centering
  \renewcommand{\thetable}{F1}
  \caption{Design space of LLM-based eligibility assessment. Properties of other systems are taken from their papers' abstracts and descriptions.}
  \label{tab:F1}
  \footnotesize
  \begin{tabularx}{\linewidth}{P{0.98}P{1.01}P{0.99}P{1.07}P{0.95}}
    \toprule
    \textbf{Property} & \textbf{Generative, criterion-level (e.g. TrialGPT)} & \textbf{Fine-tuned classifier (e.g. LLM-Match)} & \textbf{Formal / symbolic (VERDICT)} & \textbf{TrialJev pointer-head scorer} \\
    \midrule
    Output & text rationale + label & class score & verdict + formal decision trace & probability over two options \\
    Autoregressive generation & yes, per criterion & no & LLM formalizer + solver & no \\
    Per-criterion explanation & yes & not reported & yes (trace) & no \\
    Output must be parsed & yes & no & yes & no \\
    Probability usable for thresholding & indirect & yes & not reported & yes \\
    Task-specific training & no (prompting) & yes & distilled formalizer & yes (LoRA) \\
    Local / open-weight option & model-dependent & yes & model-dependent & yes \\
    Cost per decision & grows with generated tokens & one forward pass & grows with generated tokens & one forward pass \\
    \bottomrule
  \end{tabularx}
\end{table}

Published systems differ in task, granularity, data and metric, so no number in Table F2 can be set directly against ours. The table places this work among them and does not rank it.

\begin{table}[htbp]
  \centering
  \renewcommand{\thetable}{F2}
  \caption{Context from related work. Numbers are taken from the cited papers or, for the official run, from the track overview. ``Comparable?'' asks whether the number can be set against our TREC 2022 pair-level results.}
  \label{tab:F2}
  \footnotesize
  \begin{tabularx}{\linewidth}{P{0.94}P{1.14}P{1.14}P{0.98}P{0.80}}
    \toprule
    \textbf{System} & \textbf{What it does} & \textbf{Evaluation} & \textbf{Reported} & \textbf{Comparable to ours?} \\
    \midrule
    TrialJev & non-generative pointer-head pair scorer & TREC 2022, 6,975 eligible/excluded pairs, 50 patients & 80.97\% accuracy, AUROC 0.885 (Qwen3.5-27B); 81.84\%, AUROC 0.888 (Qwen3.8-27B); 70.71\%, AUROC 0.786 (0.8B, MedJev init, 3 seeds) & --- \\
    VERDICT \cite{ref13} & LLM formalization + solver, with decision traces & 363 selected TREC 2021 pairs, near-balanced & F1 0.828 (GPT-5-mini), 0.829 (distilled 7B) & same binary task family; different year, sample, balance, metric \\
    LLM-Match \cite{ref14} & fine-tuned Llama-3-8B-Instruct + RAG + classification head & TREC 2022 (also SIGIR, TREC 2021); binary label merges excluded and eligible against irrelevant & AUROC 0.8102 on TREC 2022; 0.8010 average & no, a relevance task and not eligible vs excluded \\
    TrialGPT \cite{ref1} & LLM retrieval, criterion-level matching, ranking & manual review of patient--criterion pairs; trial ranking & 87.3\% criterion-level accuracy on 1,015 reviewed pairs & different granularity; not pair-level \\
    TrialMatchAI \cite{ref9} & end-to-end retrieval + criterion reasoning & oncology patients; criterion-level classification & relevant trial in top 20 for 92\% of patients; >90\% criterion-level accuracy & different tasks and metrics \\
    Controlled LLM reasoning \cite{ref10} & retrieval + LLM reasoning & TREC 2022 ranking & best nDCG@10 0.693 (coarse); best P@10 0.730 (hybrid), different variants & full-collection ranking metrics \\
    Best official TREC 2022 run \cite{ref15} & full-collection ranking (h2oloo, frocchio\_monot5\_e) & TREC 2022 & nDCG@10 0.6125, P@10 0.5080 & ranking over the whole collection \\
    \bottomrule
  \end{tabularx}
\end{table}

The nearest published number is VERDICT's F1 of 0.83 on its own TREC 2021 subset. Our 9B to 27B models (accuracy 78.2 to 81.8\%, macro-F1 77.9 to 81.7\% for the headline entries of Table A1) are in the same range, but on all judged-relevant 2022 pairs and across years, and the differences in year, sample, construction and metric rule out a ranking. LLM-Match's AUROC of 0.81 measures relevance with excluded and eligible merged, so it cannot be set against our eligible-versus-excluded AUROC of 0.86 to 0.89. The cited studies do not provide a directly comparable pair-level eligible-versus-excluded AUROC on this TREC 2022 pool. The ranking systems answer a different question, which trials to surface from a collection, and are complementary: a retriever of that kind could feed our scorer.

\end{document}